%% file: main.tex
\documentclass{article}

\usepackage{arxiv}

\usepackage[utf8]{inputenc} 
\usepackage[T1]{fontenc}    
\usepackage{hyperref}       
\usepackage{url}            
\usepackage{booktabs}       
\usepackage{amsfonts}       
\usepackage{nicefrac}       
\usepackage{microtype}      
\usepackage{lipsum}		
\usepackage{graphicx}
\usepackage{natbib}
\usepackage{doi}

\usepackage{amsmath,amsfonts,amsthm,bm}
\usepackage{multirow}
\usepackage{booktabs}
\usepackage{subfig}
\usepackage{cleveref}
\usepackage{float}
\usepackage{makecell}

\def\vx{{\bm{x}}}
\def\vy{{\bm{y}}}
\def\vz{{\bm{z}}}
\def\vw{{\bm{w}}}
\def\vh{{\bm{h}}}
\def\vp{{\bm{p}}}

\newtheorem{theorem}{Theorem}

\title{Convection-Diffusion Equation: A Theoretically Certified Framework for Neural Networks}
\date{}
 
\author{{Tangjun Wang} \\
	Department of Mathematical Sciences\\
	Tsinghua University\\
	\texttt{wangtj20@mails.tsinghua.edu.cn} \\
	\And
	{Chenglong Bao}\thanks{Corresponding authors.} \\
	Yau mathematical sciences center\\
	Tsinghua University\\
	\texttt{clbao@mail.tsinghua.edu.cn} \\
    \And
	{Zuoqiang Shi}\footnotemark[1] \\
	Yau mathematical sciences center\\
	Tsinghua University\\
	\texttt{zqshi@tsinghua.edu.cn} \\
}

\date{}

\hypersetup{
pdftitle={A template for the arxiv style},
pdfsubject={q-bio.NC, q-bio.QM},
pdfauthor={David S.~Hippocampus, Elias D.~Striatum},
pdfkeywords={First keyword, Second keyword, More},
}

\begin{document}
\maketitle

\begin{abstract}
	\input{data/abstract}
\end{abstract}


\section{Introduction}
\input{data/introduction}

\section{Theoretical Results}
\input{data/results/theory}

\section{Experimental Results}
\input{data/results/algorithm}
\input{data/results/cora}
\input{data/results/fewshot}
\input{data/results/englandcovid}

\input{data/results/prostrate}

\section{Discussion}
\input{data/discussion}

\section{Materials and Methods}
\input{data/matmethods}

\section{Conclusion}
\input{data/conclusion}

\bibliographystyle{unsrtnat}
\bibliography{reference}

\appendix
\newpage
\begin{center}
{\Large \bf Supplementary Information}
\end{center}
\input{data/appendix}

\end{document}

%% file: data/abstract.tex
In this paper, we study the partial differential equation models of neural networks. Neural network can be viewed as a map from a simple base model to a complicate function. Based on solid analysis, we show that this map can be formulated by a convection-diffusion equation. This theoretically certified framework gives mathematical foundation and more understanding of neural networks. Moreover, based on the convection-diffusion equation model, we design a novel network structure, which incorporates diffusion mechanism into network architecture. Extensive experiments on both benchmark datasets and real-world applications validate the performance of the proposed model.

%% file: data/introduction.tex
\newcommand{\cdotspace}{{\mspace{2mu}\cdot\mspace{2mu}}}

Neural networks~(NNs) have achieved great success in many tasks, such as image classification~\cite{simonyan2015deep}, speech recognition~\cite{dahl2011context}, video analysis~\cite{bo2011object}, and action recognition~\cite{wang2016temporal}. Among the existing networks, residual networks (ResNets) are important architectures that enable the training of ultra-deep NNs and have the ability to avoid gradient vanishing~\cite{he2016deep, he2016identity}. Moreover, the idea of ResNets has motivated the development of many other NNs, including WideResNet~\cite{zagoruyko2016wide}, ResNeXt~\cite{xie2017aggregated}, and DenseNet~\cite{huang2017densely}.

In recent years, understanding ResNets from a dynamical perspective has become a promising approach~\cite{weinan2017proposal, haber2018learning, chen2018neural}. Specifically, assuming $\vx_0\in \mathbb{R}^d$ as the input of a ResNet and defining $\mathcal F$ as the mapping, the $l$-th residual block can be realized by
\begin{equation}
    \label{eq:resnet}
    \vx_{l+1}=\vx_l+\mathcal F(\vx_l,\vw_l)
\end{equation}
where $\vx_l$ and $\vx_{l+1}$ are the input and output of the residual mapping, and $\vw_l$ is the parameter of the $l$-th block that will be learned by minimizing the training loss. Let $\vx_L$ be the output of a ResNet with $L$ blocks, then the classification score is determined by $\vy = \text{softmax}(\vw\cdotspace\vx_L)$, where $\vw$ is a learnable weight of the final linear classifier.

For any $T>0$, by introducing a temporal partition $\Delta t = T/L$, the residual block represented by~\eqref{eq:resnet} can be viewed as the explicit Euler discretization with time step $\Delta t$ for the following ordinary differential equation~(ODE):
\begin{equation}
    \label{eq:ode}
    \frac{{\rm d}\vx(t)}{{\rm d}t}=v(\vx(t),t), \quad \vx(0) = \vx_0, \quad t\in[0,T],
\end{equation}
where $v(\vx(t),t)$ is a velocity field such that $v(\vx(t),t)=F(\vx(t),\vw(t))/\Delta t$. This interpretation of ResNets provides a new perspective for viewing NNs and has inspired the development of many networks. Some approaches involve applying different numerical methods to construct diverse discrete layers~\cite{larsson2017fractalnet,zhang2017polynet,lu2018beyond}, while others explore the continuous-depth model~\cite{chen2018neural,jia2019neural}.

Furthermore, the connection between ODEs and partial differential equations~(PDEs) through the well-known characteristics method has motivated the analysis of ResNets from a PDE perspective. This includes theoretical analysis~\cite{sonoda2019transport}, novel training algorithms~\cite{sun2020stochastic}, and improvements in adversarial robustness~\cite{wang2020enresnet} for NNs. Specifically, from the PDE theory, \eqref{eq:ode} is the characteristic curve of the convection equation:
\begin{equation}
    \label{eq:pde}
    \frac{\partial u}{\partial t}(\vx,t)= - v(\vx,t)\nabla u(\vx,t),\quad (\vx,t)\in\mathbb R^{d}\times[0,T].
\end{equation}

The method of characteristics tells us that, along the curve defined by~\eqref{eq:ode}, the function value $u(\vx,t)$ remains unchanged. Denote the flow map from $\vx(0)$ to $\vx(T)$ along~\eqref{eq:ode} as $\Phi$, which is the continuous form of feature extraction in ResNets. If we enforce $u(\vx,T)=f(\vx):=\text{softmax}(\vw\cdotspace\vx)$ as a linear classifier at $t=T$, then
\[u(\vx(0),0)=u(\vx(T),T)=f(\vx(T))=f\circ \Phi(\vx(0))\]
Thus at $t=0$, $u(\cdotspace,0)$ is the composition of a feature extractor and a linear classifier, which corresponds to a ResNet.

In one word, NN can be viewed as the image $u(\cdotspace,t)$ of a mapping driven by a certain PDE. In the case of ResNets, this mapping is formulated as a convection equation. A natural question is: \textit{How to bridge NN and PDE in a unified framework?}

\begin{figure*}[t]
	\centering
	\includegraphics[width=\linewidth]{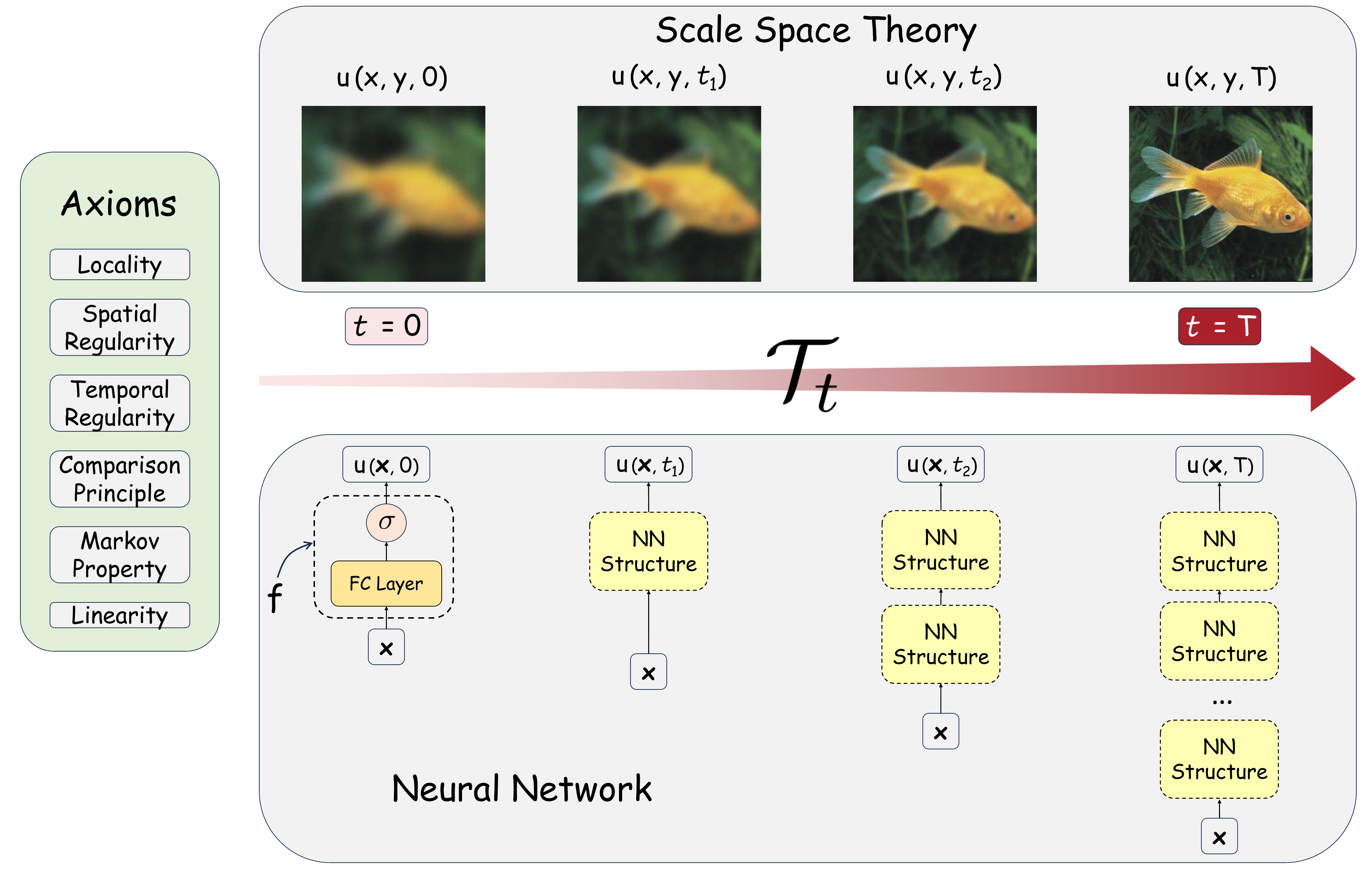}
	\caption{$\mathcal T_t$ represents the mapping from $u(\cdotspace,0)$ to $u(\cdotspace,t)$. Top block describes the evolution from a coarse image to a fine image in scale-space theory. Bottom block describes the evolution from a base classifier to a neural network.}
	\label{fig:illustration}
\end{figure*}

In this paper, we try to address the questions from a mathematical perspective. To begin, we formally define the mapping $\mathcal{T}_t$ as follows:
\[\mathcal{T}_t: f=u(\cdotspace,0) \mapsto u(\cdotspace,t), \quad t\in[0,T]\]
This operator converts a base model into a neural network. Here we choose $u(\vx,0) = f(\vx)$ because it is more intuitive to conceptualize NNs as progressing from shallow to deep in the forward direction. The connection between convection equation and ResNets remains consistent, since~\eqref{eq:pde} is reversible in time.

Defining such an evolution operator is inspired by the scale-space theory~\cite{koenderink1984structure, witkin1987scale}, which is a framework widely used in image processing, computer vision and many fields. It provides formalized theory for manipulating the image at different scales. Scale, in this context, measures the degree of smoothing, or more specifically, the size of neighborhoods of the smoothing kernel. Scale space is the family of smoothed images parameterized by the scale, from the finest image (original image) to the most coarse image. Previous works~\cite{canny1986computational,perona1990scale,alvarez1993axioms,duits2004axioms} relate scale-space theory to PDEs, where the smoothing kernels are governed by a set of axioms and satisfy a certain form of PDEs. In this paper, we aim to identify a set of criteria that the operator $\mathcal{T}_t$ should satisfy. By meeting these assumptions, we can derive the form of PDEs, and thus answer the aforementioned questions. An illustration of our approach is provided in~\Cref{fig:illustration}.

Our framework shares certain similarities with that in scale-space theory while also possessing distinct features. Both frameworks rely on several PDE-type assumptions, such as locality and regularity, because both are built upon PDEs. However, there are notable differences between the two: (1) Scale-space theory typically operates in low-dimensional spaces, such as 2D for images and 3D for movies, whereas NNs can be potentially high-dimensional. (2) NNs possess unique assumptions that are not universal in scale-space theory. On the other hand, certain assumptions from image processing, such as rotation invariance and scale invariance, cannot be directly applied to NNs. (3) The intuition behind similar assumptions may differ between the two frameworks, such as the comparison principle. We argue that our explanation of several assumptions is more natural from a NN viewpoint.

In what follows, we theoretically prove that under reasonable assumptions on $\mathcal T_t$, $u(\vx,t)=\mathcal T_t f(\vx)$ is the solution of a second order convection-diffusion PDE,
\[\frac{\partial u(\vx,t)}{\partial t}=v(\vx,t)\cdot \nabla u(\vx,t)+\sum_{i,j}\sigma_{i,j}\frac{\partial^2u }{\partial x_i\partial x_j}(\vx,t)\]
We believe that an axiomatic formulation can improve the interpretability of NNs. The theoretical result provides a unified framework which covers various existing network structures such as diffusive graph neural network, and training algorithm designed to improve robustness such as randomized smoothing. The framework also illuminates new thinking for designing networks. Specifically, we propose a new network structure called \textbf{CO}nvection d\textbf{I}ffusion \textbf{N}etworks~(COIN), which achieves state-of-the-art or competitive performance on several benchmarks as well as  novel tasks.

%% file: data/results/theory.tex
In this section, we show that under several reasonable assumptions, the sequence of operator images $u(\vx,t)=\mathcal T_t f(\vx)$ is the solution of the convection-diffusion equation. Throughout this section, we assume $\mathcal T_t$ is well defined on $C^{\infty}_b$, where $C^{\infty}_b$ is the space of bounded functions which have bounded derivatives at any order, and $\mathcal T_t f$ is a bounded continuous function. These assumptions are reasonable, as typical base classifiers $f$ are indeed bounded (between 0 and 1) and have bounded derivatives. The operator image $\mathcal T_t f$, which we hope to be a NN, is evidently bounded and continuous.

To get the expression of the operator $\mathcal T_t$, we assume it has some fundamental properties, which fall into two categories: NN-type and PDE-type. We will present the assumptions individually and provide a concise explanation of their underlying intuition. A more comprehensive discussion regarding these assumptions can be found in the Discussion section.

\subsection{NN-type assumptions}
\paragraph{[Comparison Principle]}
For all $t\geq 0$ and $f,g \in C^{\infty}_b$, if $f\leq g$, then $\mathcal T_t(f)\leq \mathcal T_t(g)$.

Suppose we are given two classifiers $f$ and $g$ such that $f(\vx) \leq g(\vx)$ for all data point $\vx \in \mathbb{R}^{d}$. Then $f\circ \Phi(\vx) \leq$ $g\circ \Phi(\vx)$ if we replace the data points $\vx$ with the extracted features $\Phi(\vx)$. Recall that for ResNet, $f\circ \Phi = \mathcal T_T(f)$, which implies $\mathcal{T}_{T}\left(f\right) \leq \mathcal{T}_{T}\left(g\right)$. Since the order-preserving property holds both at initial time step $t=0$ and final time step $t=T$, it is reasonable to make the assumption.

\paragraph{[Markov Property]} For all $s,t\geq 0$ and $t+s\leq T$, $\mathcal T_{t+s} = \mathcal T_t \circ \mathcal T_{t+s,t}$, where $\mathcal T_{t+s,t}$ denotes the flow from time $t$ to time $t+s$.

The prediction of a deep neural network is computed using forward propagation, i.e. the network uses output of former layer as input of current layer. Thus, for a NN model, it's natural that the output of a NN can be deduced from the output of intermediate $l$-th layer without any information depending upon the original data point $\vx$ and output of $m$-th layer ($m<l$). Regarding the evolution of operator $\mathcal T_t$ as stacking layers in the neural network, we should require that $\mathcal T_{t+s}$ can be computed from $\mathcal T_t$ for any $s\geq 0$, and $T_0$ is of course the identity.

\paragraph{[Linearity]} For any $f,g \in C^{\infty}_b$, and real constants $\beta_1,\beta_2$, we have 
\[\mathcal T_{t}(\beta_1 f+\beta_2 g) = \beta_1\mathcal T_{t}(f)+\beta_2\mathcal T_{t}(g)\]
if $C$ is a constant function, then $\mathcal T_t(C) = C$.

Linearity is also an intrinsic property of deep neural networks. Notice that we are not referring to a single NN's output v.s. input linearity, which is obviously wrong because of the activation function. Rather, we are stating that two different NN with the same feature extractor can be merged in to a new NN with a new classifier composed with the shared extractor, i.e.
\[(\beta_1f+\beta_2g) \circ \Phi=\beta_1 f \circ \Phi + \beta_2 g \circ \Phi \]
This is linearity at $t=T$, and for $t=0$ it is trivial. For the latter part, we can hope that a constant base model always produces constant values with evolution.

\subsection{PDE-type assumptions}

\paragraph{[Locality]} For all fixed $\vx$, if $f,g \in C_b^{\infty}$ satisfy $D^{\alpha}f(\vx) = D^{\alpha}g(\vx)$ for all $|\alpha|\geq 0$, where $D^{\alpha}f$ denotes the $\alpha$-order derivative of $f$, then
\[\lim_{t\rightarrow 0^+}\frac{(\mathcal T_t(f)-\mathcal T_t(g))(\vx)}{t}=0\]

First of all, we need an assumption to ensure the existence of a differential equation. If two classifiers $f$ and $g$ have the same derivatives of any order at some point, then we should assume same evolution at this point when $t$ is small. If we unrigorously define $\partial T_t(f)/\partial t = \left(\mathcal T_t(f)-f\right)/t$ when $t\rightarrow 0^+$ (or infinitesimal generator in our proof), then $\partial T_t(f)/\partial t$ should equal to $\partial T_t(g)/\partial t$. Thus, we give the locality assumption concerning the local character of the operator $\mathcal T_t$ for $t$ small. 

\paragraph{[Spatial Regularity]} There exists a positive constant $C$ depending on $f$ such that 
\[\|\mathcal T_t(\tau_{\vh} f)-\tau_{\vh} (\mathcal T_t f)\|_{L^{\infty}}\leq Cht\]
for all $f\in C^{\infty}_b, \vh \in \mathbb R^d, t\geq 0$, where $(\tau_{\vh} f)(\vx)=f(\vx+\vh)$ and $\|\vh\|_2=h$.

Regularity is an essential component in PDE theory. Thus, when considering PDE-type assumptions on $\mathcal T_{t}$, it is necessary to study its regularity. We separate the regularity requirements into spatial and temporal. Spatial regularity implies that the addition of a perturbation $\vh$, whether applied to the base model $f$ or the evolved model $T_t f$, should result in minimal differences. One can relate it to the well-known translation invariance in image processing, but our assumption is weaker, as we allow small difference rather than require strict equivalence.

\paragraph{[Temporal Regularity]}
For all $t,s,t+s\in [0,T]$ and all $f\in C^{\infty}_b$, there exist a constant $C\geq 0$ depending on $f$ such that
\begin{align*}
    \|\mathcal T_{t+s,s}(f)-f\|_{L^{\infty}}& \leq Ct\\
    \|\mathcal T_{t+s,s}(f)-\mathcal T_t(f)\|_{L^{\infty}} & \leq Cst
\end{align*}

Temporal regularity requires that in any small time interval, the evolution process will not be rapid, because we want a smooth operator $\mathcal T_t$ in time.

\vbox{}

Finally, combine all the assumptions on $\mathcal T_{t}$, we can derive the following theorem, emphasizing that the output value of neural network $T_t(f)$ with time evolution satisfies a convection-diffusion equation,

\begin{theorem}
    \label{thm:main}
    Under the above assumptions, there exists Lipschitz continuous function $v:\mathbb{R}^d\times [0,T]\to \mathbb{R}^d$ and Lipschitz continuous positive function $\sigma:\mathbb{R}^d\times [0,T]\to \mathbb R^{d\times d}$ such that for any bounded and uniformly continuous base classifier $f(\vx)$, $u(\vx,t) = \mathcal T_t(f)(\vx)$ is the unique solution of the following convection-diffusion equation:
    \begin{equation}
        \label{eq:main}
    	\begin{cases}
    	\frac{\partial u(\vx,t)}{\partial t}=v(\vx,t)\cdot \nabla u(\vx,t)+\sum_{i,j}\sigma_{i,j}\frac{\partial^2u }{\partial x_i\partial x_j}(\vx,t),\\
    	u(\vx,0)=f(\vx),
    	\end{cases}
    \end{equation}
    where $\vx\in \mathbb R^{d}, t\in [0,T]$.
    Here $\sigma_{i,j}$ is the $i,j$-th element of matrix function $\sigma(\vx,t)$. 
\end{theorem}

We will provide the proof of \Cref{thm:main} in the Supplementary Information.

%% file: data/results/algorithm.tex
\subsection{Convection Diffusion Network}
\begin{figure}[hbtp]
    \hfill
	\subfloat[$u(\vx, 0)$]{\includegraphics[page=1, width=0.2\linewidth]{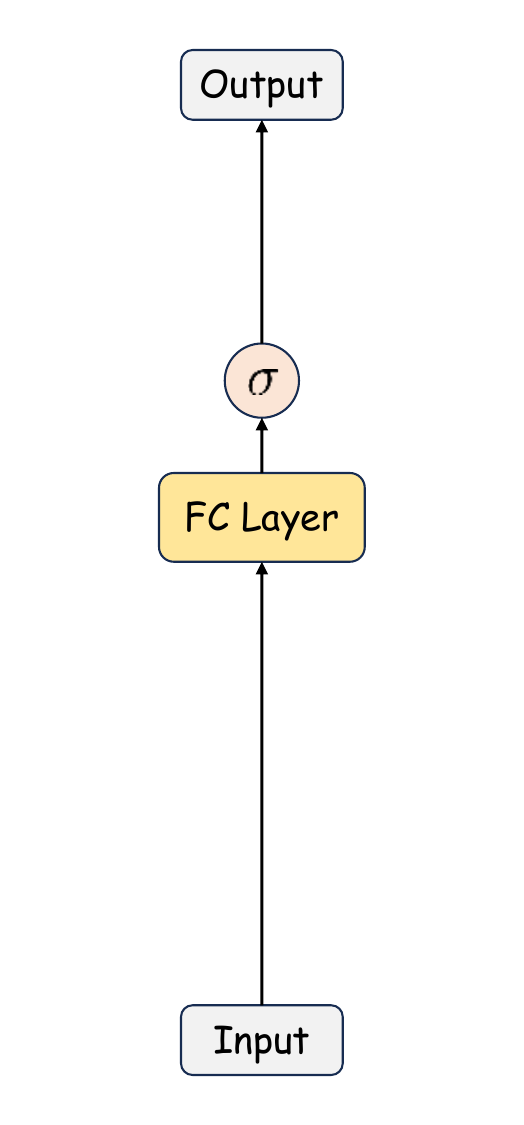}}
    \hfill
    \subfloat[$u(\vx, T-1)$]{\includegraphics[page=2, width=0.2\linewidth]{figures/network_structure.pdf}}
    \hfill
    \subfloat[$u(\vx, T)$]{\includegraphics[page=3, width=0.2\linewidth]{figures/network_structure.pdf}}
    \hfill
\end{figure}

To validate the effectiveness of our axiomatized PDE model, we have designed a novel network structure called~\textbf{CO}nvection d\textbf{I}ffusion \textbf{N}etworks~(COIN), which incorporates diffusion layers after the ResNet architecture. 

The proposed algorithm is a straightforward split scheme of~\eqref{eq:main}. We divide the convection-diffusion equation into two parts, namely, the convection part and the diffusion part,
\begin{align*}
    \begin{cases}
    \frac{\partial u(\vx,t)}{\partial t}= v(\vx,t) \cdot \nabla u(\vx,t), &\quad  t\in [0,T-1]\\
    \frac{\partial u(\vx,t)}{\partial t} = \sigma^2 \Delta u(\vx,t), &\quad t\in[T-1,T]\\
    u(\vx,0) = f(\vx)
    \end{cases}
\end{align*}
Here we focus on the isotropic models, i.e., the diffusion term is $\sigma^2\Delta u$. The time step $T-1$ serves as a pseudo time step for understanding and does not have a practical impact. As mentioned in the introduction, the forward propagation of ResNets can be viewed as the convection equation. Therefore, we can use a ResNet to simulate the convection part from 0 to $T-1$. In our implementation, we use a shallow two-layer fully connected network with a residual connection to represent the ResNet.

To handle the diffusion part, we model the data samples as nodes on a graph, allowing us to discretize the Laplacian term using the graph Laplacian. In this context, a graph represents a discretization of the domain $\mathbb{R}^n$ into a finite space, where a continuous function vector $u$ is defined. Given a graph $\mathcal{G}=({\cal V},{\cal E})$, where $\mathcal{V}=\{\vx_i\}^N_{i=1}$ denotes the set of $N$ vertices and ${\cal E}=\{w_{ij}\}_{i,j=1}^N$ describes the relationship between nodes $\vx_i$ and $\vx_j$, we can compute the graph Laplacian as follows:
\begin{equation*}
    \Delta u(\vx,t)=-Lu(\vx,t)=-\sum_{j=1}^N w_{ij} \left(u(\vx_i,t)-u(\vx_j,t)\right)
\end{equation*}
The weight $w_{ij}$ can be either given or pre-computed, depending on the specific task. $L=D-W$ is the graph Laplacian matrix, where $D=\text{diag}(d_i)$ is the diagonal matrix with entries $d_i=\sum_{j=1}^N w_{ij}$. Then, by applying the forward Euler scheme to discretize the derivative with respect to time $t$, we obtain the following expression:
\begin{equation}
    u^{k+1}_i=u^k_i - \sigma^2 \sum_{j} w_{ij} (u_i^k-u_j^k), \quad k=0, 1, \cdots, K-1
    \label{eq:diffusion_layer}
\end{equation}
where $u_i^k$ represents the value of $u$ on node $\vx_i$ at time step $t_k$. The initial time step $t_0$ is set to $T-1$, which corresponds to the output of the ResNet, and the final time step $t_K=T$ corresponds to the final output of COIN. \eqref{eq:diffusion_layer} is referred to as a diffusion layer. In our implementation, we often stack multiple diffusion layers ($K>1$) because the value of $\sigma^2$ cannot be too large due to stability concerns~\cite{wang2024diffusion}. Consequently, we use the forward Euler scheme to discretize the diffusion term, allowing us to reach the desired diffusion strength.

To demonstrate that our implementation is derived from the PDE, rather than ODE, perspective, we should point out that diffusion is imposed on the network output value $u$, rather than intermediate value $\vx$. We achieve such idea by incorporating the activation function (such as softmax for multi-class classification or sigmoid for binary classification) in the output layer of ResNet represented by $u(\vx, T-1)$. Consequently, when computing the final loss function, it is unnecessary to add an additional activation function after $u(\vx, T)$.

Last but not least, we want to emphasize that our COIN model is only one of the many possible approaches of modeling the convection-diffusion PDE. Other methods may include introducing a regularization term that enforces NNs to obey the PDE, similar to PINN~\cite{raissi2019physics}. We are looking forward to exploring other paths in the future work.

%% file: data/results/cora.tex
\subsection{Graph Node Classification}

We have performed tests on our COIN model for semi-supervised node classification problems in graph. We present the results for the well-established citation network benchmarks, namely Cora, Citeseer, and Pubmed. These datasets consist of citation networks where nodes represent publications, edges represent citation links, and features are represented by sparse bag-of-words vectors. The dataset statistics are provided in Materials and Methods. Each node (publication) is categorized into a class. The objective of the graph node classification task is to predict the class of test nodes, given only a limited number of labeled training nodes.

To ensure a reliable comparison, we follow the methodology of Shchur et al.~\cite{shchur2019pitfalls} instead of using the fixed Planetoid split~\cite{yang2016revisiting}. We conduct 100 random train-val-test splits, with each split involving 20 random neural network initialization. We report the average accuracy and standard variation across these splits. For each dataset, we adopt the approach of Shchur et al.~\cite{shchur2019pitfalls} and select 20 data points per class for the training set, 30 data points per class for the validation set, and the remaining points for the test set.

\begin{table}[hbtp]
\centering
\caption{Performance comparison on graph node classification tasks. Reported results are average accuracy $\pm$ standard variation from 100*20 experiments. Methods marked with $^*$ indicate that their results are averaged across 40 random splits with 10 random initialization each, due to excessive time consumption per task.}
\label{tab:cora}
\begin{tabular}{c@{\hspace{1.2\tabcolsep}}cc@{\hspace{1.2\tabcolsep}}c@{\hspace{1.2\tabcolsep}}c}
	\toprule
	& Method & Cora & Citeseer & Pubmed \\
	\midrule
	\multirow{6}{*}{Classic} & MLP & 57.4 $\pm$ 2.1 & 59.9 $\pm$ 2.2  & 70.0 $\pm$ 2.0 \\
	& GCN~\cite{kipf2017semi} & 81.6 $\pm$ 1.1 & 72.1 $\pm$ 1.6 & 79.0 $\pm$ 2.1 \\
	& GraphSAGE~\cite{hamilton2017inductive} & 79.3 $\pm$ 1.4 & 71.7 $\pm$ 1.6 & 76.1 $\pm$ 2.0 \\
	& GAT~\cite{velickovic2018graph} & 80.8 $\pm$ 1.3 & 71.6 $\pm$ 1.7 & 78.7 $\pm$ 2.1 \\
	& SGC~\cite{wu2019simplifying} & 80.5 $\pm$ 1.3 & 73.9 $\pm$ 1.4 & 77.2 $\pm$ 2.6 \\
	& APPNP~\cite{gasteiger2019predict} & \textbf{82.7} $\pm$ 1.1 & 73.3 $\pm$ 1.5 & 80.6 $\pm$ 1.8 \\
	\midrule
	\multirow{2}{*}{ODE} & CGNN$^*$~\cite{xhonneux2020continuous} & 82.5 $\pm$ 1.0 & 73.0 $\pm$ 1.6 & 80.5 $\pm$ 2.2 \\
	& GCDE~\cite{poli2021graph} & 80.0 $\pm$ 1.5 & 72.1 $\pm$ 1.6 & 76.0 $\pm$ 3.9 \\
	\midrule
	\multirow{6}{*}{Diffusion} & GDC~\cite{gasteiger2019diffusion} & 81.6 $\pm$ 1.3 & 72.2 $\pm$ 2.6 & 79.0 $\pm$ 2.0 \\
	& GraphHeat$^*$ ~\cite{xu2019graph} & 81.4 $\pm$ 1.2 & 73.5 $\pm$ 1.5 & 78.4 $\pm$ 2.1 \\
	& DGC~\cite{wang2021dissecting} & 81.4 $\pm$ 1.2 & 75.0 $\pm$ 1.9 & 78.2 $\pm$ 2.1 \\
	& Difformer~\cite{wu2023difformer} & 82.0 $\pm$ 2.3 & 71.9 $\pm$ 1.7 & 74.8 $\pm$ 4.5 \\
	& GRAND~\cite{chamberlain2021grand} & 82.5 $\pm$ 1.4 & 73.7 $\pm$ 1.7 & 78.8 $\pm$ 1.8 \\
	& Diff-ResNet~\cite{wang2024diffusion} & 82.1 $\pm$ 1.1 & 74.6 $\pm$ 1.8 & 80.1 $\pm$ 2.0 \\
	\midrule
	& COIN & 82.2 $\pm$ 1.2 & \textbf{75.8} $\pm$ 1.3 & \textbf{81.1} $\pm$ 1.9 \\
	\bottomrule
\end{tabular}
\end{table}

We compare our method with graph learning methods from three categories that are closely related to ours: classic methods, ODE-based methods, and methods that also include diffusion. Some methods may belong to more than one category, e.g. GRAND. In this case, we pick its main contribution as the category. We have re-implemented all the aforementioned methods using their official code available online and compared their performance with ours under the same experimental settings. During the testing of these methods, we have used the recommended parameters provided in the paper or the Github repository. A comparison between our re-implemented results and reported results is provided in the Supplementary Information. For detailed training settings, please refer to Materials and Methods.

The experimental results are presented in \Cref{tab:cora}. Our COIN outperforms state-of-the-art approaches in terms of accuracy on Citeseer and Pubmed, while achieving comparable results on Cora.

\begin{figure}[hbtp]
	\subfloat[Cora $K=20$]{\includegraphics[width=0.33\linewidth]{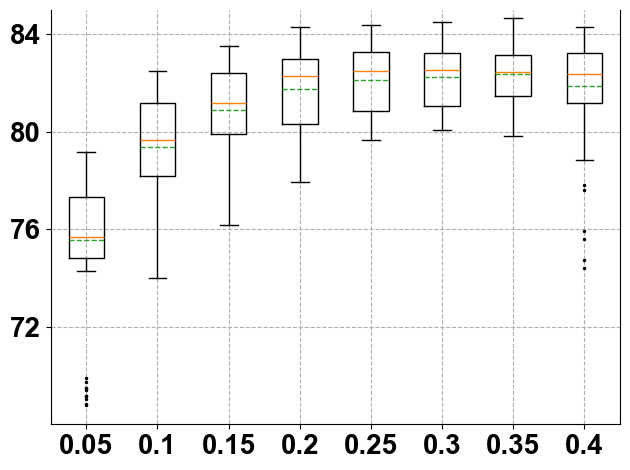}}
	\hfill
	\subfloat[Citeseer $K=20$]{\includegraphics[width=0.33\linewidth]{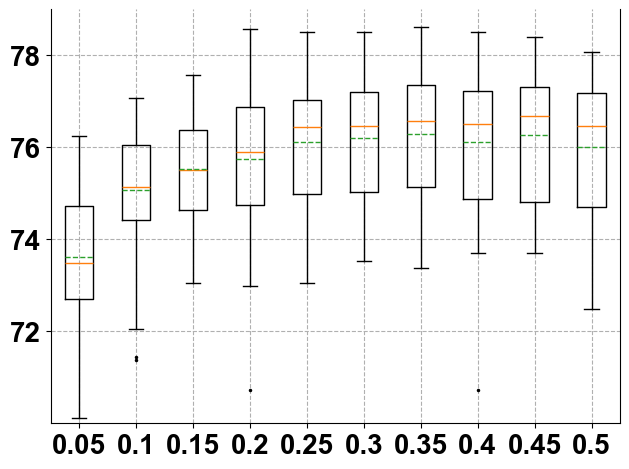}}
	\hfill
	\subfloat[Pubmed $K=20$]{\includegraphics[width=0.33\linewidth]{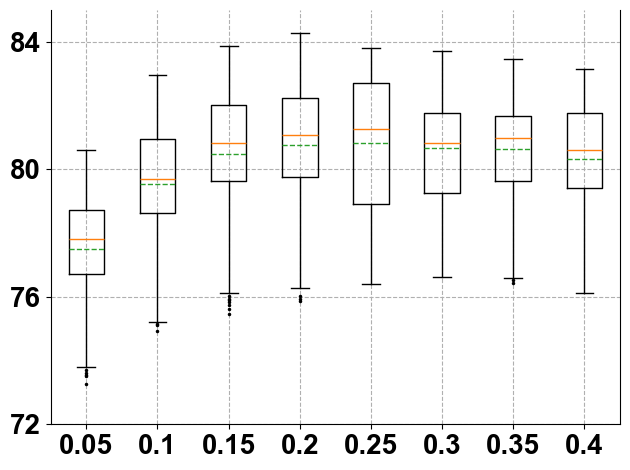}}
		
	\subfloat[Cora $K=40$]{\includegraphics[width=0.33\linewidth]{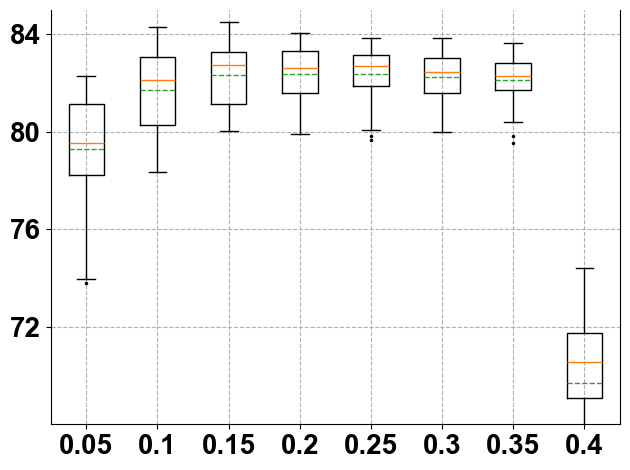}}
	\hfill
	\subfloat[Citeseer $K=40$]{\includegraphics[width=0.33\linewidth]{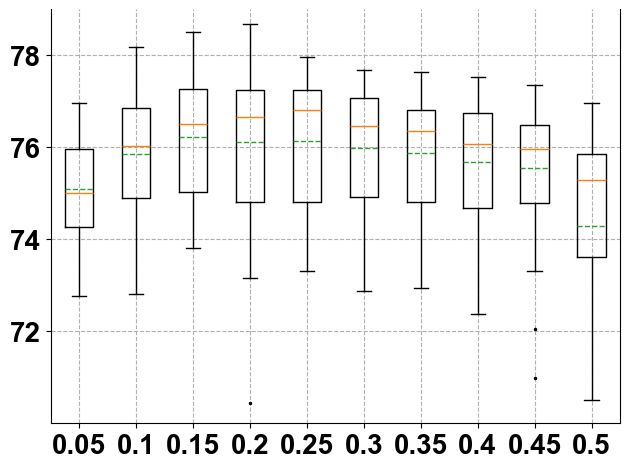}}
	\hfill
	\subfloat[Pubmed $K=40$]{\includegraphics[width=0.33\linewidth]{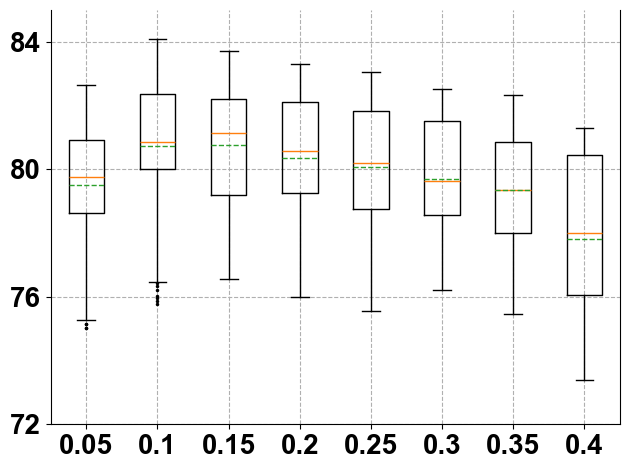}}

    \caption{Accuracy boxplots of COIN with different diffusion strength. x-axis represents $\sigma^2$, y-axis represents accuracy(\%). The orange solid line represents median. The green dashed line represents mean. The lower and upper hinges correspond to the 1st and 3rd quartiles, the whisker corresponds to the minimum or maximum values no further than 1.5 $\times$ inter-quartile range from the hinge. Data beyond the end of the whiskers are outlying points that are plotted individually.}
    \label{fig:cora_ablation}
\end{figure}

Furthermore, we study the effect of diffusion strength on the network performance. We fix the number of diffusion layers at either $K=20$ or $K=40$ and vary the diffusion strength of each layer $\sigma^2$. The accuracy is averaged over 10 random train-validation-test splits, with each split involving 10 random neural network initializations, thus potentially yielding slightly different results compared to Table \ref{tab:cora}. We plot the boxplot of 100 experimental results in~\Cref{fig:cora_ablation}.

From the results, two findings emerge. First, the performance gradually increases and then decreases with the total diffusion strength $K\sigma^2$, which indicates the existence of an optimal strength. Notably, the network achieves comparable performance within a fairly broad range of diffusion strength. Second, for a given total strength $k\sigma^2$, the layer number $K$ does not have a significant effect. For example, the performance with $K=20$ and $\sigma^2=0.3$ is comparable to that with $K=40$ and $\sigma^2=0.15$.

%% file: data/results/fewshot.tex
\subsection{Few-shot learning}

The effectiveness of deep learning methods is strongly influenced by the availability of a substantial number of training examples. However, collecting such data requires significant labor and is often unfeasible in many domains due to the privacy or safety issues. To alleviate the reliance on training data, there has been a growing interest in few-shot learning methods~\cite{fei2006one, vinyals2016matching} in recent years. See~\cite{wang2020generalizing} for a comprehensive review. Formally, the few-shot learning tasks are defined as follows. Given a novel dataset $\mathbb{X}_{\text{novel}}= \mathbb{X}_{\text{s}}\cup\mathbb{X}_{\text{q}}$, where $\mathbb{X}_{\text{s}} = \{(x_i,y_i)\}_{i=1}^{N_1}$ is the support set with label information and $\mathbb{X}_{\text{q}} = \{x_j\}_{j=1}^{N_2}$ is the query set without labels, the goal of few-shot learning is to find the labels of points in the query set when the size of support set $|N_1|$ is very small. Along with the novel dataset $\mathbb{X}_{\text{novel}}$, there exists a base dataset $\mathbb{X}_{\text{base}}$, where all samples are provided with label information. To prevent information leak, $\mathbb{X}_{\text{base}}$ and $\mathbb{X}_{\text{novel}}$ contain data points from distinct classes. The base dataset can be utilized for various purposes, such as data augmentation in transfer learning, episodic training in meta-learning, or backbone training in embedding learning. Among the different few-shot learning method, we employ embedding learning, which aims to map each sample into a latent space such that similar samples are close while dissimilar samples are far away. The embedding function is parameterized by a deep neural network (backbone) pretrained on $\mathbb{X}_{\text{base}}$. During the few-shot learning tasks on $\mathbb{X}_{\text{novel}}$, the pretrained embedding function remains fixed without any further fine-tuning.

\begin{table}[hbtp]
\caption{1-shot average accuracy (in \%) and 95\% confidence interval in \textit{mini}ImageNet, \textit{tiered}ImageNet and CUB with backbone WRN. Re-implemented results using public official code with our pretrained backbone are marked with \dag.}
\label{tab:1-shot}
\centering
\begin{tabular}{lccc}
\toprule
Methods & \textit{mini}ImageNet & \textit{tiered}ImageNet & CUB \\
\hline
Qiao \cite{qiao2018few} & 59.60 $\pm$ 0.41 & - & - \\
LEO \cite{rusu2018meta} & 61.76 $\pm$ 0.08 & 66.33 $\pm$ 0.05 &-\\
ProtoNet \cite{snell2017prototypical} & 62.60 $\pm$ 0.20 & - & -\\
CC+rot \cite{gidaris2019boosting} & 62.93 $\pm$ 0.45 & 70.53 $\pm$ 0.51 &-\\
MatchingNet \cite{vinyals2016matching} & 64.03 $\pm$ 0.20 & - & -\\
FEAT \cite{ye2020fewshot} & 65.10 $\pm$ 0.20 & 70.41 $\pm$ 0.23 &-\\
Transductive \cite{dhillon2019baseline} & 65.73 $\pm$ 0.68 & 73.34 $\pm$ 0.71 &-\\
BD-CSPN \cite{liu2020prototype} & 70.31 $\pm$ 0.93 & 78.74 $\pm$ 0.95 &-\\
PT+NCM \cite{hu2021leveraging} & 65.35 $\pm$ 0.20 & 69.96 $\pm$ 0.22 & 80.57 $\pm$ 0.20 \\
SimpleShot \cite{wang2019simpleshot}$^{\dag}$ & 65.20 $\pm$ 0.20 & 71.49 $\pm$ 0.23 & 78.62 $\pm$ 0.19 \\
LaplacianShot \cite{ziko2020laplacian}$^{\dag}$ & 72.90 $\pm$ 0.23 & 78.79 $\pm$ 0.25 & 87.70 $\pm$ 0.18 \\
EPNet \cite{rodriguez2020embedding}$^{\dag}$ & 67.09 $\pm$ 0.21 & 73.20 $\pm$ 0.23 & 80.88 $\pm$ 0.20 \\
Diff-ResNet \cite{wang2024diffusion}$^{\dag}$ & 73.47 $\pm$ 0.23 & 79.74 $\pm$ 0.25 & 87.74 $\pm$ 0.19 \\
COIN & \textbf{74.85} $\pm$ 0.24 & \textbf{80.66} $\pm$ 0.25 & \textbf{89.20} $\pm$ 0.18 \\
\bottomrule
\end{tabular}
\end{table}

We conduct experiments on three benchmarks for few-shot image classification: \textit{mini}ImageNet, \textit{tiered}ImageNet and CUB. The \textit{mini}ImageNet and \textit{tiered}ImageNet are both subsets of the larger ILSVRC-12 dataset~\cite{russakovsky2015imagenet}, with 100 classes and 608 classes respectively. The \textit{mini}ImageNet contains 100 classes and is split into 64 base classes, 16 validation classes, and 20 novel classes. The \textit{tiered}ImageNet contains 608 classes and is split into 351 base classes, 97 validation classes, and 160 novel classes. CUB-200-2011~\cite{wah2011caltech} is a fine-grained image classification dataset with 200 classes. It is split into 100 base classes, 50 validation classes, and 50 novel classes. The dataset split is standard as in previous papers~\cite{wang2019simpleshot,ziko2020laplacian}. All images are resized to $84 \times 84$, following~\cite{vinyals2016matching}. 

The most common way to build a task is called an $N$-way-$K$-shot task~\cite{vinyals2016matching}, where $N$ classes are sampled from $\mathbb{X}_{\text{novel}}$ and only $K$ (e.g., 1 or 5) labeled samples are provided for each class. Following standard evaluation protocol~\cite{wang2019simpleshot,ziko2020laplacian}, we randomly sample 10000 5-way-1-shot and 5-way-5-shot classification tasks, and report the average accuracy and corresponding 95\% confidence interval. 

We choose two widely used networks, ResNet-18~\cite{he2016deep} and WRN-28-10~\cite{zagoruyko2016wide} as our backbone. The backbone training process is in general similar to that in SimpleShot~\cite{wang2019simpleshot} and LaplacianShot~\cite{ziko2020laplacian}, but details are slightly different. See the Materials and Methods for details. The training process follows a standard pipeline in supervised learning and does not involve any meta-learning or episodic-training strategy. As a result, we obtain an embedding function that maps the original data points to $\mathbb{R}^M$, where $M=512$ for ResNet-18 and $M=640$ for WRN-28-10. For fair comparison, we also employ techniques used in \cite{wang2019simpleshot,ziko2020laplacian}, including centering and normalization and cross-domain shift, to transform the embedded features. The details of these techniques are explained in the Materials and Methods. 

During each few-shot task, we train a COIN using $M$-dimensional features extracted from both the support set and the query set. To apply diffusion, we require a weight matrix that captures the similarity between data points. The weight $w_{ij}$ is computed using a Gaussian kernel based on the Euclidean distance between $x_i$ and $x_j$.

The results of 1-shot tasks with WRN-28-10 as backbone are reported in \Cref{tab:1-shot}. Results for 5-shot tasks and ResNet-18 as backbone are available in the Supplementary Information. In \Cref{tab:1-shot}, the results for comparison are collected from \cite{wang2019simpleshot,ziko2020laplacian}. Across all datasets with different backbone architectures, our COIN consistently achieves the highest classification accuracy. Our method outperforms the state-of-the-art by an average margin of more than 1\%, showcasing its effectiveness in scenarios with extremely limited training data.

%% file: data/results/englandcovid.tex
\subsection{COVID-19 case prediction with missing data}

Our axiomatic framework can also be applied to predict the reported case for COVID-19 in scenarios with missing data. During a pandemic, accurate prediction of the infection spread is of utmost importance to enable governments to take timely and proactive measures. Timely and accurate predictions enable governments to make informed decisions, optimize healthcare resource allocation, and implement effective measures to suppress the spread of the virus. Recent studies have addressed the challenge of predicting pandemics using deep learning methods~\cite{zeroual2020deep, chimmula2020time, pal2020neural, hu2020artificial}. One promising approach is to use graph neural network~\cite{kapoor2020examining, panagopoulos2021transfer, gao2021stan, fritz2022combining}, where regions are considered as nodes, and the interaction between nodes is captured through human mobility data (i.e., the number of people moving from one place to another within a given period).

We conduct our experiment using the England COVID-19 dataset sourced from the PyTorch Geometric Temporal open-source library~\cite{rozemberczki2021pytorch}. This dataset comprises daily reported COVID-19 cases in 129 regions of England known as NUTS3 regions, spanning from 3rd March to 12th May (61 days). The dataset is structured as a collection of graph snapshots, where each snapshot represents a specific day. Within each graph, the nodes correspond to the 129 regions in England. The node features capture the number of COVID-19 cases reported in each region over the past 8 days, while the objective is to predict the number of cases in each node for the following day. The graphs are directed and weighted, with edge weights indicating the daily volume of people moving from one region to another. These weights are derived from the Facebook Data For Good disease prevention maps and the official UK government website. Importantly, the graph snapshots are dynamic, meaning that different snapshots exhibit variations in terms of node features, edge weights, and prediction targets. To ensure consistent analysis, we normalize the reported cases for both node features and targets, setting their mean to zero and their variance to one. We divide the snapshots into training, validation, and test sets using a 2:2:6 ratio chronologically.

To address the challenge of missing data, we apply a masking strategy when constructing the dataset, which randomly hides a portion of the reported cases with a probability of 0.9. Notice that a reported case can be used both as part of node feature and as target. When missing data occurs in the node features, we substitute it with a value of zero. As for the missing data in the target, we treat it as NaN (Not a Number). This means that on the specific day (graph snapshot), the region with missing target data does not contribute to the neural network training. We achieve this by masking out the loss when the target is NaN during training. This setting effectively simulates the scenario of missing data that occurs in real-world situations, closely resembling the challenges encountered in practical applications. However, during testing, we make an exception and use the true target values for more accurate evaluation. Nevertheless, any missing data in the node feature is still treated as zero.

\begin{table}[hbtp]
    \centering
    \caption{Average Mean Squared Error on England COVID-19 dataset with missing data.}
    \label{tab:covid}
	\begin{tabular}{cc}
    	\toprule
    	Method & MSE \\
    	\midrule
    	All-Zero & 0.8197 \\
    	LR & 0.8344 $\pm$ 0.0769 \\
    	MLP & 0.7777 $\pm$ 0.0203 \\
    	GCN & 0.7607 $\pm$ 0.0185 \\
        DCRNN~\cite{li2017diffusion} & 0.8213 $\pm$ 0.0663 \\
        MPNN-LSTM~\cite{panagopoulos2021transfer} & 0.9384 $\pm$ 0.0633 \\
        GConvGRU~\cite{seo2018structured} & 0.8153 $\pm$ 0.0435\\
        A3TGCN~\cite{bai2021a3t} & 0.7797 $\pm$ 0.0362\\
        EvolveGCN~\cite{pareja2020evolvegcn} & 0.8237 $\pm$ 0.0856 \\ 
    	COIN & \textbf{0.7168} $\pm$ 0.0187 \\
    	\bottomrule
	\end{tabular}
\end{table}

We compared our COIN model with several methods using Mean Squared Error (MSE) on the test set as the evaluation metric. To provide a fair benchmark, we establish a baseline where we simply predict all-zero values. Since the dataset has been normalized to zero mean, this baseline serves as a reference point. Additionally, we implemented Logistic Regression (LR), Multi-Layer Perceptron (MLP), and Graph Convolutional Network (GCN) to compare against our COIN model. These models represent traditional and widely-used approaches in the field. Furthermore, we explore various methods that leverage techniques from recurrent neural networks (RNN) to incorporate temporal information. These methods not only consider spatial information but also incorporate temporal information.

We randomly mask out the reported cases using 10 random seeds, each with 10 random neural network initialization. The results are presented in \Cref{tab:covid}. In our experimental setup, where a significant portion of the data is missing, methods that incorporate temporal information struggle to generalize and some perform even worse than the all-zero baseline. We speculate that the misleading effect of missing data on network predictions is amplified when employing temporal information within an RNN structure.

Conversely, non-RNN methods such as MLP and GCN outperform the aforementioned approaches. The superiority of GCN over MLP can be attributed to the inclusion of graph information, specifically the daily movement of people, which plays a crucial role in pandemic prediction, especially when only a limited amount of valid data from the past several days is available in our setting. By leveraging information from neighboring nodes, the neural network can make more accurate predictions.

Nevertheless, our COIN model surpasses all other methods by a significant margin, with a notable 74.4\% improvement over GCN when compared to the all-zero baseline.

%% file: data/results/prostrate.tex
\subsection{Prostrate cancer classification}

Over the past decade, significant advancements in molecular profiling technologies have enabled the collection of genomic, transcriptional and additional features from cancer patients. This wealth of molecular profiling data, combined with clinical annotations, has greatly contributed to the identification of numerous genes, pathways, and complexes associated with lethal cancers. However, in the case of prostrate cancer, uncovering the relationship between these molecular features and disease prediction remains a major biological and clinical challenge~\cite{robinson2015integrative,abida2019genomic,elmarakeby2021biologically}. Developing a highly accurate predictive model is essential for the early detection of preclinical prostate cancer and holds significant potential for wider application in various cancer types.

The dataset utilized in this study is derived from the work of Armenia et al.~\cite{armenia2018long}, where they collected and analyzed genomic profiles from 1,013 patients diagnosed with prostate cancer (680 primary and 333 metastatic). These profiles are generated using a unified computational pipeline to ensure consistent derivation of somatic alterations. Following~\cite{elmarakeby2021biologically}, patient features are aggregated at the gene level, resulting in a total of 9,229 genes. Each gene is encoded with binary values (0 or 1) to represent somatic mutation, copy number amplification, and copy number deletion. Consequently, each patient is characterized by a feature vector of dimension 27,687. The dataset is split into a training set of size 20, a validation set of size 100 and the rest as the test set. The objective is to predict the cancer state of patients in the test set as either primary or metastatic.

We compare our COIN model with several machine learning methods and a deep neural network approach. Machine learning methods include logistic regression~(LR), support vector machine~(SVM), random forest and adaptive boosting~(AdaBoost). The deep neural network approach, P-NET~\cite{elmarakeby2021biologically}, incorporates curated biological pathways to build a pathway-aware multi-layered hierarchical network. In this network, each neuron represents a gene, and the connections between neurons correspond to biological pathways. As a result, P-NET is a sparse network with 6 layers and a total of 71,009 parameters. 

Following~\cite{elmarakeby2021biologically}, we adopt the same first layer structure from~\cite{elmarakeby2021biologically} in Multi-Layer Perceptron~(MLP), ResNet and COIN in~\Cref{tab:prostrate}. Specifically, each neuron is connected to exactly three nodes in the input feature representing somatic mutation, copy number amplification, and copy number deletion of one gene. To ensure that the performance improvement of our COIN over P-NET does not result from an increase in parameters, the hidden dimension of MLP, ResNet and COIN is set to 3. Consequently, the total number of parameters is 64610 for MLP, and 73840 for ResNet and COIN, which is comparable to that of the P-NET model. Similar to few-shot learning, the weight is computed using a Gaussian kernel based on the Euclidean distance between patient features. However, instead of using the raw 27,687 dimension feature, we pretrain a ResNet and select the 9,229 dimension output after the first layer as the feature for computing distance.

\begin{table}[hbtp]
    \centering
    \caption{Average accuracy, ROC-AUC and AUPRC on prostrate classification task.}
    \label{tab:prostrate}
	\begin{tabular}{cccc}
    	\toprule
    	Method & Accuracy(\%) & ROC-AUC & AUPRC \\
    	\midrule
    	LR & 67.90 & 0.5884 & 0.4498 \\
        SVM & 67.01 & 0.6484 & 0.4370 \\
        Random forest & 67.14 & 0.7307 & 0.5150 \\
        AdaBoost & 76.75 & 0.7094 & 0.5967 \\
        MLP & 73.39 & 0.6540 & 0.5498 \\
    	P-NET~\cite{elmarakeby2021biologically} & 74.77 & 0.7720 & 0.6548 \\
        ResNet & 78.23 & 0.7635 & 0.7100 \\
    	COIN & \textbf{80.35} & \textbf{0.8001} & \textbf{0.7598} \\
    	\bottomrule
	\end{tabular}
\end{table}

We conduct a comparison of the accuracy, area under the receiver operating characteristic curve (ROC-AUC), and area under the precision-recall curve (AUPRC) for these methods. ROC and PRC shows the true positive rate v.s. false positive rate, and precision v.s. recall, respectively, at different classification thresholds. The results, presented in Table \ref{tab:prostrate}, are averaged over 10 random train-validation-test splits and 10 random initializations for each split. Notably, our COIN consistently achieves the highest performance across all evaluation metrics. Compared with ResNet, out COIN adds several diffusion layers motivated by PDE framework and further improves the classification performance. These results underscore the effectiveness of our axiomatic framework in accurately classifying prostate cancer state.

%% file: data/discussion.tex
\subsection{Examples under framework}
Under the convection-diffusion framework, we can give interpretation to several regularization mechanisms that are designed to improve robustness, including Gaussian noise injection~\cite{wang2020enresnet,liu2020does}, ResNet with stochastic dropping out the hidden state of residual block~\cite{srivastava2014dropout, sun2020stochastic} and randomized smoothing~\cite{cohen2019certified,li2019certified,salman2019provably}. We can also interpret several graph neural networks, including diffusive ones~\cite{eliasof2021pdegcn,chamberlain2021grand,wang2021dissecting,wang2024diffusion} and convective ones~\cite{xhonneux2020continuous,poli2021graph}, as a specific example under our unified framework.

\paragraph{Gaussian noise injection}
Gaussian noise injection is an effective regularization mechanism for a NN model. For a vanilla ResNet with $L$ residual mapping, the $l$-th residual mapping with Gaussian noise injected can be written as
\[\vx_{l+1}=\vx_l + \mathcal F(\vx_l,\vw_l)+ a \mathcal{N}(0,\mathbf I)\]
where the parameter $a$ is a noise coefficient. Using the same temporal partition as in the introduction, and let $a=\sigma\sqrt{\Delta t}$, this noise injection can be viewed as the approximation of the following continuous dynamic using the Euler-Maruyama method,
\begin{equation}
    \label{eq:ito}
    \mathrm{d}\vx(t)=v(\vx(t),t)\mathrm{d}t+\sigma \mathrm{d}\mathbf B(t)
\end{equation}
where $\mathbf B(t)$ is a Brownian motion. The output of $L$-th residual mapping is the state of I$\hat t$o process \eqref{eq:ito} at terminal time $T$. So, an ensemble prediction over multiple networks with shared parameters and random Gaussian noise can be written as a conditional expectation,
\begin{equation}
    \label{eq:expectation}
   \hat y=\mathbb E({\rm softmax}(\vw \vx(T))|\vx(0)=\vx_0).
\end{equation}
According to Feynman-Kac formula~\cite{mao2007stochastic}, \eqref{eq:expectation} is known to solve the following convection-diffusion equation
\begin{equation*}
\begin{cases}
	\frac{\partial u(\vx,t)}{\partial t}=v(\vx,t)\cdot \nabla u(\vx,t)+ \frac{1}{2}\sigma^2 \Delta u(\vx,t),\quad t\in [0,T]\\
	u(\vx,0)={\rm softmax}(\vw \vx).		
\end{cases}
\end{equation*}

\paragraph{Dropout of Hidden Units}
Consider the case that we disable every hidden units independently from a Bernoulli distribution $\mathcal B(1,p)$ with $p\in(0,1)$ in each residual mapping
\begin{align*}
    \vx_{l+1} & = \vx_l+\mathcal F(\vx_l,\vw_l)\odot \frac{\vz_l}{p}\\
    & = \vx_l+\mathcal F(\vx_l,\vw_l)+\mathcal F(\vx_l,\vw_l)\odot (\frac{\vz_l}{p}-\textbf{I})
\end{align*}
where $\vz_l\sim\mathcal B(1,p)$ namely $\mathbb P(\vz_n=0)=1-p$, $\mathbb P(\vz_n=1)=p$ and $\odot$ indicates the Hadamard product. If the number of the ensemble is large enough, according to Central Limit Theorem, we have
\[\mathcal F(\vx_l,\vw_l)\odot (\frac{\vz_l}{p}-\textbf{I})\approx\mathcal F(\vx_l,\vw_l)\odot\mathcal N(0,\frac{1-p}{p})\]
The similar way with Gaussian noise injection, the ensemble prediction $\hat y$ can be viewed as the solution $u(\vx,T)$ of an convection-diffusion equation with diffusion term
\[\frac{1-p}{2p}\sum_{i}(v^Tv)_{i,i}\frac{\partial^2u}{\partial x_i^2}(\vx,t)\]
Compared with adding Gaussian noise, which is an isotropic model, dropout of hidden units is an anisotropic model, because the noise introduced by dropout is related to the output of previous layer. In fact, similar to dropout, shake-shake regularization~\cite{gastaldi2017shake,huang2018stochastic} and ResNet with stochastic depth~\cite{huang2016deep} can also be interpreted by our convection-diffusion equation model.

\paragraph{Randomized Smoothing}
Consider transforming a trained classifier into a new smoothed classifier by adding Gaussian noise to the input during inference. If we denote the trained classifier by $h(\vx)$ and denote the new smoothed classifier by $g(\vx)$. Then $h(\vx)$ and $g(\vx)$ have the following relation:
\[g(\vx)=\frac{1}{N} \sum^N_{i=1} h(\vx+\bm{\varepsilon}_i) \approx \mathbb E_{\bm{\varepsilon} \sim \mathcal N(0,\sigma^2I)}[h(\vx+\bm{\varepsilon})]\]
where $\bm{\varepsilon}_i\sim \mathcal N(0,\sigma^2I)$. According to Feynman-Kac formula, $g(\vx)$ can be viewed as the solution of the following PDE
\begin{equation*}
    \begin{cases}
    \frac{\partial u(\vx,t)}{\partial t}=\frac{1}{2}\sigma^2\Delta u,\quad t\in [0,1]\\
    u(\vx,0)=h(\vx).
    \end{cases}
\end{equation*}
Especially, when $h(\vx)$ is a ResNet, the smoothed classifier $g(\vx)=u(\vx,T+1)$ can be expressed as
\begin{equation*}
	\begin{cases}
         \frac{\partial u(\vx,t)}{\partial t}=v(\vx,t)\cdot \nabla u(\vx,t),\quad t\in [0,T]\\
		\frac{\partial u(\vx,t)}{\partial t}=\frac{1}{2}\sigma^2\Delta u, \hspace{46pt} t\in [T,T+1]\\
		u(\vx,0)={\rm softmax}(\vw\vx).
	\end{cases}
\end{equation*}
We extend the terminal time $T$ to $T+1$ to emphasize that randomized smoothing is a post-processing step.

\paragraph{Diffusive Graph Neural Network}
Some models, e.g. PDE-GCN~\cite{eliasof2021pdegcn}, GRAND~\cite{chamberlain2021grand}, DGC~\cite{wang2021dissecting}, introduce diffusion in graph neural network, which corresponds to the diffusion part in our convection-diffusion framework. Diff-ResNet~\cite{wang2024diffusion} contains both convection and diffusion, but its formulation is derived from ODE perspective. Thus the diffusion is applied on the features $\vx$, rather than on the value $u(\vx,t)$.

\paragraph{Convective Graph Neural Network}
There are some methods, including CGNN~\cite{xhonneux2020continuous} and GCDE~\cite{poli2021graph}, which models the forward propagation from the perspective of ODE,
\[\frac{{\rm d}\vx(t)}{{\rm d}t}=v(\vx(t),t)\]
As stated in the introduction, it can be viewed as the characteristics of the convective PDE,
\[\frac{\partial u(\vx,t)}{\partial t}=-v(\vx,t)\cdot \nabla u(\vx,t),\quad t\in [0,T]\]
Nonetheless, these methods belong to the ODE category, ignoring the diffusion term during forward propagation.

\subsection{More explanation on assumptions}
We compare our assumptions with the corresponding assumptions in the scale space theory, providing their common ground and difference below.

\paragraph{[Comparison Principle]}
In scale-space theory, especially for the transformation on grey-scale images, the comparison principle means that if one grey-scale image is everywhere brighter than another image, this ordering should be preserved along with the smoothing of the original picture. However, when the goal is to find the edges or depth map in the image, the comparison principle is no longer valid. However, from our NN perspective, comparison principle is natural and always valid, since the ordering between classifiers should be the same, no matter the input is raw data point or extracted feature.

\paragraph{[Markov Property]}
Markov Property corresponds to \textit{Recursivity}, or slightly weaker \textit{Causality}, in scale space theory. It is natural that a coarser analysis of the original picture can be inferred from a finer one without any reliance on the original picture itself. It is also natural in NN due to the feed-forward structure of network design.

\paragraph{[Linearity]}
Linearity is not a common assumption in the field of scale space theory, as there exists both linear processes~\cite{koenderink1984structure,canny1986computational} and nonlinear processes~\cite{perona1990scale}, depending on the filters. However, it is an inherent property of NN, as discussed in the theoretical results. Such linear combination is widely used in practice known as ensemble methods~\cite{zhou2012ensemble}. It is a popular method in machine learning which combines many weak classifiers and add them to obtain a strong classifier. 

\paragraph{[Locality]}
Locality describes the local characterization of the operator, and is crucial for both scale space theory and ours, as they both use PDE to describe the evolution of the operator. Using the assumptions other than Locality, we may prove the existence of an infinitesimal generator, $\left(\mathcal T_t(f)-f\right)/t$. Then, the meaning of locality is that if two functions have the same derivatives at some point, then they have the same infinitesimal generator at this point.

\paragraph{[Spatial Regularity]}
The most relevant axiom in scale-space theory is translation invariance, which means
\[\mathcal T_t(\tau_{\vh} f)=\tau_{\vh} (\mathcal T_t f)\]
It is stronger than our assumption as it requires absolute equivalence. The concept of translation invariance stems from the idea that there should be no prior knowledge about the specific location of a feature in an image. In convolutional neural networks~\cite{lecun1998gradient}, translation invariance is achieved by a combination of convolutional layers and pooling layers. For example, a cat is recognized as a cat regardless of whether it appears in the top or bottom half of an image. However, we do not intend to manually design the neural network structure to strictly enforce this level of invariance. Instead, we relax the constraint to allow for some variations.

Spatial regularity may also be beneficial for adversarial robustness. NNs have been shown to be vulnerable to some well-designed input samples, which are called adversarial samples~\cite{goodfellow2014explaining,kurakin2016adversarial}. These adversarial samples are produced by adding carefully hand-crafted perturbations to the inputs of the targeted model. Although these perturbations are imperceptible to human eyes, they can fool NNs to make wrong prediction. In some sense, the existence of these adversarial examples is due to spatial unstability of NNs. We hope the new model $\mathcal T_t(f)$ to be spatially stable by adding spatial regularity.

\paragraph{[Temporal Regularity]}
Temporal regularity is used in both scale-space theory and in our NN framework, which ensures the existence of PDE. The first inequality states a natural assumption about continuity. When $s=0$, it reduces to
\[\|\mathcal T_t(f)-f\|_{L^{\infty}} \leq Ct\]
which means that the change should be small when the evolution time is short. It is related to stiffness in the field of numerical solution of differential equations. A stiff equation generally means that there is rapid variation in the solution, and thus we need extremely small steps when numerically solving the equation. Stiffness is not a desirable property of differential equations, and thus we require temporal regularity in the framework. The second inequality is a natural extension for time origin from $t=0$ to $t\geq 0$.

\subsection{Extra Comments}

Why do we want to use an axiomatic framework? Why do we say it can improve the interpretability of NNs? Some may question this, arguing that since the analytic expression of solutions to convection-diffusion equations is intractable, the framework has nothing to do with interpretability. However, it is important to clarify that we are referring to the interpretability of the model itself, rather than the interpretability of the model results.

An instructive would be Maxwell's equations, which serve as the foundation of classical electromagnetism. These equations are derived from several fundamental laws in physics, such as Gauss's law and Faraday's law. They help scientists understand the underlying principles governing electricity, magnetism, and light. Although it is generally not possible to solve Maxwell's equations analytically, this does not diminish the interpretability provided by the equations. Similarly, in our framework, the solution to the convection-diffusion model is not the primary concern; rather, it is the assumptions made and the intuition behind those assumptions that truly matter.

\begin{figure}[htb]
\centering
    \includegraphics[width=0.5\linewidth]{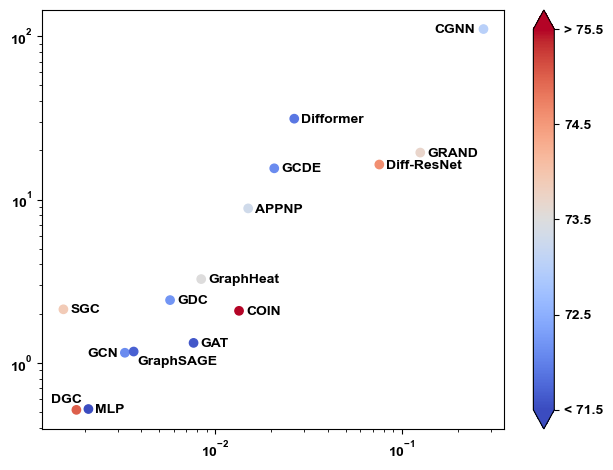}
    
    \caption{Time complexity on Citeseer dataset. The x-axis represents average time (seconds) per training epoch, and the y-axis represents average convergence time (seconds) per task, both in log-scale. The color bar measures the average accuracy of each method.}
    \label{fig:time_complexity}
\end{figure}

\begin{figure}[hbtp]
\centering
    \includegraphics[width=0.5\linewidth]{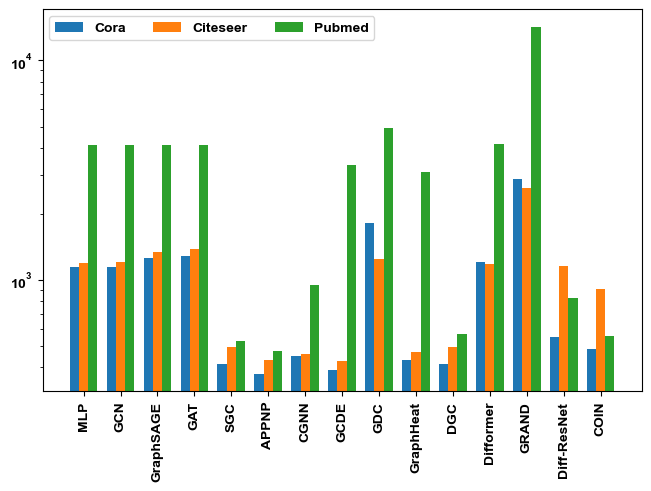}
    \caption{Space complexity shown in bar plot. The y-axis represents the allocated GPU memory (MB) in log-scale.}
    \label{fig:space_complexity}
\end{figure}

Regarding the time and space complexity of our algorithm, we provide the average time per training epoch, average convergence time per task, and allocated GPU memory of our COIN, compared to other methods, on graph node classification tasks. The experiments are conducted on a single NVIDIA GeForce RTX 3090. It is worth noting that we have used the same early-stopping criterion for all methods, ensuring a fair comparison in terms of average convergence time. Our method incorporates diffusion on the output $u=f(\vx)\in \mathbb{R}^c$, where $c$ represents the number of classes. As the class number is significantly smaller than the feature dimension (e.g., 7 vs. 1433 for Cora dataset), our method exhibits low time and space complexity, even with multiple diffusion layers. As shown in~\Cref{fig:time_complexity}, our method demonstrates significantly reduced per-epoch time and convergence time compared to certain methods in the ODE and Diffusion categories, such as CGNN and GRAND, while achieving superior results. The time complexity results for Cora and Citeseer are provided in the Supplementary Information. In~\Cref{fig:space_complexity}, even with multiple diffusion layers ($K=20$ in graph node classification tasks), COIN does not occupy much memory.

Currently, the weight in the adjacency matrix is either given or pre-computed, which fix the correlation between data samples. Nonetheless, it is possible to introduce attention mechanism to dynamically adjust the pairwise weight. By learning the attention parameter, we may model more complex relationship, and we leave as future work.

%% file: data/matmethods.tex
\subsection{Graph Node Classification}
For all datasets, we treat the graph as undirected and only consider the largest connected component.
\begin{table}[hbtp]
	\caption{Graph Node Classification Dataset Statistics.}
	\centering
	\begin{tabular}{lccccc}
		\toprule
		Dataset & Node & Edge & Class & Feature Dim & Label Rate \\ \midrule
		Cora    & 2485  & 5069  & 7       & 1433        & 0.057     \\
		Citeseer    & 2120  & 3679  & 6       & 3703        & 0.056     \\
		Pubmed    & 19717  & 44324  & 3       & 500        & 0.003     \\ \bottomrule
	\end{tabular}
\end{table}

We follow the normalization technique in GCN~\cite{kipf2017semi}: the adjacent matrix is first added with a self-loop, and then symmetrically normalized. The feature vectors are row normalized.

We use the Adam optimizer with learning rate 0.01. Weight decay is 5e-4 for Cora and Pubmed, 0.01 for Citeseer. Dropout is not used in this experiment. The early stopping criteria is the validation loss does not decrease and the validation accuracy does not increase for 50 epochs. Parameters are chosen based on the accuracy on the validation set. The number of diffusion layers $K$ is fixed to be 20 for all datasets. The strength for each layer $\sigma^2$ is 0.35 for Cora, 0.4 for Citeseer, and 0.3 for Pubmed.

\subsection{Few-shot learning}
We train the backbone on the base classes using cross-entropy loss with label smoothing factor of 0.1, SGD optimizer, standard data augmentation and a mini-batch size of 256 to train all models. The model is trained for $T=100$ epochs for \textit{mini}ImageNet and \textit{tiered}ImageNet, and $T=400$ epochs for CUB due to its small size. We use a multi-step scheduler, which decays the learning rate by 0.1 at $0.5T$ and $0.75T$. We evaluate the nearest-prototype classification accuracy on the validation set and obtain the best model.

Following previous works\cite{wang2019simpleshot, ziko2020laplacian}, two additional feature transformation skills are used to enhance the performance. (1) Centering and Normalization: $x=x-\bar{x} \quad \text{then} \quad x=x/\|x\|_2,~\forall x\in \mathbb{X}_s\cup\mathbb{X}_q$,
where $\bar{x}$ is the base class average. (2) Cross-Domain Shift: $x=x+\Delta,~\forall x\in \mathbb{X}_q$, where
$\Delta=\frac{1}{|\mathbb{X}_s|}\sum_{\mathbb{X}_s}x-\frac{1}{|\mathbb{X}_q|}\sum_{\mathbb{X}_q}x$
is the difference between the mean of features within the support set and the mean of features within the query set. 

Weight is calculated by $w_{ij}= \exp(-\|x_i-x_j\|_2^2/\sigma(x_i)^2)$, where $\sigma(x_i) = k$ means $\sigma$ is chosen to be the $k$-th closest distance from a specific point $x_i$ , so it varies with points. We choose $n_{\mathrm{top}}=8$, which truncates weight matrix to the 8-th nearest neighbor, and $\sigma=4$. Weight is symmetrically normalized. The diffusion step size $\sigma^2$ is fixed to be 0.5 for all tasks. The number of diffusion layers number $K$ varies with tasks: for 1-shot learning, $K=10$ for all datasets; for 5-shot learning, $K=4$ for \textit{mini}ImageNet and CUB, $K=2$ for \textit{tiered}ImageNet. The optimizer is SGD with initial learning rate $0.1$, momentum $=0.9$ and weight\_decay $=1e$-4. We train $T=100$ epochs. We use a multi-step scheduler, which decays the learning rate by 0.1 at $0.5T$ and $0.75T$.

\subsection{COVID-19 case prediction with missing data}
For all methods, we select ELU~\cite{clevert2016fast} as the activation function due to its slightly better performance. When normalizing the adjacency matrix for PDE-ResNet, we exclude the self-loop by setting the diagonal elements to zero. We also experiment with removing the self-loop for other methods, but find that doing so negatively affects their performance. Hence, we only employ this technique for our PDE-ResNet.

We use the Adam optimizer with a learning rate of 0.01 and weight decay of 5e-4 to train our network for 100 epochs. We choose a hidden dimension of 16 for MLP, ResNet and our PDE-ResNet. The dropout rate is set to 0.5 for all other methods, while for PDE-ResNet, it is set to 0.25, because we apply dropout after each diffusion layer. The selection of the dropout rate is based on testing the MSE on the validation set using a grid search. In our model architecture, we used $K=10$ layers with $\sigma^2=0.5$, which aligns with the settings in few-shot learning.

\subsection{Prostrate cancer classification}
We use the SGD optimizer with a initial learning rate of 1.0 and weight decay of 5e-4 to train our PDE-ResNet for $T=300$ epochs. We use a multi-step scheduler, which decays the learning rate by 0.1 at $0.5T$ and $0.75T$. When calculating the weight, we first pretrain a ResNet for 300 epochs with the same optimizer as above. Output after the first layer is used as the feature to calculate weight for each patient. Weight is calculated in the same way as in few-shot learning, and the parameters are $n_{\mathrm{top}}=40$, $\sigma=20$. Diffusion parameters are $K=40$ and $\sigma^2=0.2$.

\subsection{Data, Materials, and Software Availability}
Code and data for reproducing results in the paper has been deposited in \href{https://github.com/shwangtangjun/COIN}{https://github.com/shwangtangjun/COIN}.

%% file: data/conclusion.tex
Motivated by the scale-space theory, we theoretically prove that the evolution from a base classifier to neural networks can be modeled by a convection-diffusion equation. Based on the theoretical results, we develop a novel network structure and verify its effectiveness through extensive experiments. We are aware that modeling the convection-diffusion equation through introducing diffusion mechanism is one of the many possible approaches, and we are looking forward to explore other paths in the future work.

%% file: data/appendix.tex
\section{Proof of Theorem 1}
\begin{proof}
Following the techniques in~\cite{alvarez1993axioms}, we set
\[\delta_{t,s}(f)=\frac{\mathcal T_t(f)-\mathcal T_s(f)}{t-s}, \quad \delta_t(f)=\delta_{t,0}(f)\]
The proof of theorem mainly consists of two steps. First we will prove that $\delta_{t}(f)$ converges to a limit as $t\rightarrow 0$, which we call an infinitesimal generator. Then, we verify that the generator satisfies a second-order convection-diffusion equation.

First of all, we describe some basic properties of $\delta_t(f)$. From \textbf{[Temporal Regularity]}, we know $\delta_t(f)$ is uniformly bounded,
\[ \|\delta_t(f)\|_{L^{\infty}} =\left\|\frac{\mathcal T_t(f)-f}{t}\right\|_{L^{\infty}} \leq C \]
Also, it is obvious that \textbf{[Linearity]} is preserved for $\delta_t(f)$,
\[\delta_t(\beta_1 f+\beta_2 g) = \beta_1 \delta_t(f)+\beta_2 \delta_t(g)\]
Additionally, $\delta_t(f)$ is Lipschitz continuous on $\mathbb{R}^d$, uniformly for $t\in(0,1]$ and $f\in C^{\infty}_b$. Indeed, let $\vh \in \mathbb R^d, \|\vh\|_2=h$,
\[\|\tau_{\vh}(\delta_t(f))-\delta_t(f)\|_{L^{\infty}} \leq \|\tau_{\vh}(\delta_t(f))-\delta_t(\tau_{\vh}f)\|_{L^{\infty}} + \|\delta_t(\tau_{\vh}f)-\delta_t(f)\|_{L^{\infty}}\]
The first term can be bounded using \textbf{[Spatial Regularity]},
\[\|\tau_{\vh}(\delta_t(f))-\delta_t(\tau_{\vh}f)\|_{L^{\infty}} = \left\|\frac{\tau_{\vh} (\mathcal T_t f)-\mathcal T_t(\tau_{\vh} f)}{t}\right\|_{L^{\infty}}\leq Ch\]
The second term can be bounded using the fact that $f\in C^{\infty}_b$. We may write $\tau_{\vh}f=f+hg$ for some $g\in C^{\infty}_b$ depending on $\vh$, then using linearity and uniform boundedness,
\[\|\delta_t(\tau_{\vh}f)-\delta_t(f)\|_{L^{\infty}}=\|\delta_t(f+hg)-\delta_t(f)\|_{L^{\infty}}=h\|\delta_t(g)\|_{L^{\infty}}\leq Ch\]
Lastly, since $f\leq g + \|f-g\|_{L^{\infty}}$, we can use \textbf{[Comparison Principle]} and \textbf{[Linearity]} to get
\[\mathcal T_{t}(f)\leq \mathcal T_{t}(g+\|f-g\|_{L^{\infty}}) = \mathcal T_{t}(g) +\|f-g\|_{L^{\infty}}\]
Thus for any $f,g \in C^{\infty}_b$\footnote{By continuity, $T_{t}$ can be extended as a mapping from $B U C\left(\mathbb{R}^{d}\right)$, the space of bounded, uniformly continuous functions on $\mathbb{R}^{d}$ into itself. By density, \eqref{eq:a1_contraction} still hold for $f, g$ in $B U C\left(\mathbb{R}^{d}\right)$. The extension will be used in \eqref{eq:a1_estimate_2}.} and $t\geq 0$,
\begin{equation}
	\label{eq:a1_contraction}
    \|\mathcal T_{t}(f)-\mathcal T_{t}(g)\|_{L^{\infty}}\leq \|f-g\|_{L^{\infty}}
\end{equation}
Now we are ready to prove the main theorem.

\subsection{Existence of infinitesimal generator}
First we want to prove: 
\begin{equation}
	\label{eq:a1_estimate}
    \|\delta_{t+s,t}(f)-\delta_s(f)\|_{L^{\infty}}\leq m(t)
\end{equation}
where $m(t)$ is some continuous, nonnegative, nondecreasing function such that $m(0)=0$, and $m(t)$ depends only on the bounds of derivatives of $f$.

Since $\delta_s(f)$ not necessarily belongs to $C^{\infty}_b$, we mollify $\delta_s(f)$ by introducing a standard mollifier $K\geq 0$ satisfying $\int_{\mathbb R^d}K \mathrm{d}\vy=1$, $K\in C_0^{\infty}(\mathbb R^d)$ and $K_{\varepsilon}=\varepsilon^{-d}K(\cdot/\varepsilon)$. Using the Lipschitz continuity of $\delta_s(f)$, we can obtain that for all $\varepsilon>0$, there exist a positive constant $C_1$ depending only on the derivatives of $f$, such that
\begin{equation}
	\label{eq:a1_estimate_1}
    \|\delta_s(f)\ast K_{\varepsilon}-\delta_s(f)\|_{L^\infty}\leq C_1\varepsilon.
\end{equation}
where $\ast$ denote the convolution. Because of \textbf{[Markov Property]}, \textbf{[Temporal Regularity]} and \eqref{eq:a1_contraction}, we have
\begin{eqnarray}
	&&\left\|\mathcal T_{t+s}(f)-\mathcal T_t\circ \mathcal T_s(f)\right\|_{L^{\infty}} = \left\|\mathcal T_t\circ \mathcal T_{t+s,t}(f)-\mathcal T_t\circ \mathcal T_s(f)\right\|_{L^{\infty}} \label{eq:a1_estimate_2}\\
	&\leq& \left\|\mathcal T_{t+s,t}(f)-\mathcal T_s(f)\right\|_{L^{\infty}}\leq Cst \nonumber
\end{eqnarray}

By \eqref{eq:a1_contraction} and \eqref{eq:a1_estimate_1}, we have
\begin{eqnarray}
	&&\left\|\mathcal T_t\circ \mathcal T_s(f)-\mathcal T_t(f+s\delta_s(f)*K_{\varepsilon})\right\|_{L^{\infty}} \leq \left\|\mathcal T_s(f) \label{eq:a1_estimate_3} -(f+s\delta_s(f)*K_{\varepsilon})\right\|_{L^{\infty}}\\
	&=&s\|\delta_s(f)\ast K_{\varepsilon}-\delta_s(f)\|_{L^\infty}\leq C_1\varepsilon s \nonumber
\end{eqnarray}
By \textbf{[Linearity]} and \textbf{[Temporal Regularity]}, since $\delta_s(f)*K_{\varepsilon}\in C^{\infty}_b$, we have
\begin{eqnarray}
	&& \left\|\mathcal T_t(f+s\delta_s(f)*K_{\varepsilon})-(\mathcal T_t(f)+s\delta_s(f)*K_{\varepsilon})\right\|_{L^{\infty}} \label{eq:a1_estimate_4} \\
	&=& s\left\|\mathcal T_t(\delta_s(f)*K_{\varepsilon})-\delta_s(f)*K_{\varepsilon}\right\|_{L^{\infty}} \leq C_{\varepsilon}st \nonumber
\end{eqnarray}

for some positive constant $C_{\varepsilon}$ depending only on $\varepsilon$. Combining \eqref{eq:a1_estimate_1}, \eqref{eq:a1_estimate_2}, \eqref{eq:a1_estimate_3} and \eqref{eq:a1_estimate_4}, we finally deduce that
\[\|\delta_{s+t,t}(f)-\delta_s(f)\|_{L^\infty}\leq 2C_1\varepsilon+C_{\varepsilon}t+Ct\]
By setting $m(t)=\inf_{\varepsilon\in(0,1]}(2C_1\varepsilon+C_{\varepsilon}t)+Ct$ we can get desired estimate \eqref{eq:a1_estimate}.

Now we give a Cauchy estimate for $\delta_s(f)$. We will prove that
\[\|\delta_t(f)-\delta_h(f)\|_{L^{\infty}}\leq 2\frac{C_0r}{t}+m(t)\quad {\rm{where}}\ r = t - Nh, N=\left[\frac{t}{h}\right]\]
Notice that
\[\delta_t(f)=\frac{Nh}{t}\delta_{Nh}(f)+\frac{r}{t}\delta_{Nh+r,Nh}(f)\]
Using \eqref{eq:a1_estimate} with $s = r$, $t = Nh$ we have
\begin{equation}
	\label{eq:a1_use_estimate_1}
	\|\delta_{Nh+r}(f) - \frac{Nh}{Nh+r}\delta_{Nh}(f)-\frac{r}{Nh+r}\delta_r(f)\|_{L^{\infty}}\leq \frac{r}{Nh+r}m(Nh)
\end{equation}
Again, notice that
\[\delta_{Nh}(f)=\frac{(N-1)h}{Nh}\delta_{(N-1)h}(f)+\frac{h}{Nh}\delta_{Nh,(N-1)h}(f)\]
Using \eqref{eq:a1_estimate} with $s = h$, $t = (N-1)h$, we have
\begin{equation}
	\label{eq:a1_use_estimate_2}
	\|\delta_{Nh}(f)-\frac{N-1}{N}\delta_{(N-1)h}(f)-\frac{1}{N}\delta_h(f)\|_{L^{\infty}}\leq \frac{1}{N}m((N-1)h)
\end{equation}
Combining \eqref{eq:a1_use_estimate_1} and \eqref{eq:a1_use_estimate_2}, we obtain
\[\|\delta_t(f)-(N-1)\frac{h}{t}\delta_{(N-1)h}(f)-\frac{h}{t}\delta_h(f)-\frac{r}{t}\delta_r(f)\|_{L^{\infty}}\leq \frac{r}{t}m(Nh)+\frac{h}{t}m((N-1)h)\]
Reiterating the procedure, we obtain that after $(N-1)$ steps,
\[\|\delta_t(f)-\frac{Nh}{t}\delta_h(f)-\frac{r}{t}\delta_r(f)\|_{\infty}\leq \frac{r}{t}m(Nh)+\frac{h}{t}\sum_{j=1}^{N-1}m(jh)\]
Since $m(t)$ is nondecreasing and $\delta_t(f)$ is uniformly bounded, we have
\begin{eqnarray*}
	\|\delta_t(f)-\delta_h(f)\|_{L^{\infty}} & \leq &\|\delta_t(f)-\frac{Nh}{t}\delta_h(f)-\frac{r}{t}\delta_r(f)\|_{L^{\infty}}+\frac{r}{t}\|\delta_h(f)-\delta_r(f)\|_{L^{\infty}}\\
	& \leq &\frac{r}{t}m(Nh) + \frac{(N-1)h}{t}m(t) + \frac{r}{t}\left(\|\delta_h(f)\|_{L^{\infty}} + \|\delta_r(f)\|_{L^{\infty}}\right)\\
	&\leq &m(t)+2\frac{C_0 r}{t}
\end{eqnarray*}
Since $\delta_t(f)$ is uniformly bounded and Lipschitz continuous, We can pick $h_n$ going to $0$ and $\delta_{h_n}(f)$ converges uniformly on compact sets to a bounded Lipschitz function on $\mathbb{R}^d$, which we denote by $A[f]$ (the infinitesimal generator). Then using the Cauchy estimate we have derived, we have
\[\lim_{n\to\infty}\|\delta_t(f)-\delta_{h_n}(f)\|_{L^{\infty}}\leq \lim_{n\to\infty}m(t)+2\frac{C_0r}{t}\]
which implies
\[\|\delta_t(f)-A[f]\|_{L^{\infty}} \leq m(t)\]
So $\delta_t(f)$ converges uniformly to $A[f]$ when $t$ goes to $0$. Similarly, there exist an operator $A_t$ such that $\delta_{s,t}(f)$ converges uniformly to $A_t[f]$ when $s$ goes to $t$.

\subsection{Second-order convection-diffusion equation}
Let $f,g\in C^{\infty}_{b}$ and satisfy
$f(\bm{0})=g(\bm{0})=0$ (if not equal to 0, we replace $f(\vx),g(\vx)$ by $f(\vx)-f(\bm{0}),g(\vx)-g(\bm{0})$), $Df(\bm{0})=Dg(\bm{0})=\vp \in \mathbb R^{d}$, $D^{2}f(\bm{0})=D^{2}g(\bm{0})=\bm{A} \in \mathbb R^{d\times d}$. We are first going to show that $A[f](\bm{0})=A[g](\bm{0})$. 

Introduce $f^{\varepsilon}=f+\varepsilon \|\vx\|_2^{2} \in C_b^{\infty}$. Using Taylor formula, there exist a positive constant $c$ such that for $\|\vx\|_2\leq c\varepsilon$ we have $f^{\varepsilon}\geq g$. Let $w\in C^{\infty}_{b}(\mathbb R^{d})$ be a bump function satisfying 
\begin{equation*}
\begin{cases}
w(\vx)=1\,\,\,&\|\vx\|_2\leq c/2\\
0 \leq w(\vx) \leq 1\,\,\,&c/2<\|\vx\|_2<c \\
w(\vx)=0\,\,\,&\|\vx\|_2\geq c
\end{cases}
\end{equation*}
and $w_{\varepsilon}(\vx)=w(\vx/\varepsilon)$. Finally we introduce $\bar{f}^{\varepsilon}=w_{\varepsilon}f^{\varepsilon}+(1-w_{\varepsilon})g$ so that $f^{\varepsilon}_{0}\geq g$ on the whole domain $\mathbb R^{d}$. Then because of \textbf{[Comparison Principle]}, $\mathcal T_{t}(\bar{f}^{\varepsilon})\geq \mathcal T_{t}(g)$. Since $\bar{f}^{\varepsilon}(\bm{0})=f^{\varepsilon}(\bm{0})=f(\bm{0})=g(\bm{0})$, we can get $A[\bar{f}](\bm{0})\geq A[g](\bm{0})$. Because there exists a neighborhood of $\bm{0}$ that $\bar{f}^{\varepsilon} = f^{\varepsilon}$, we have $D^{\alpha}\bar{f}^{\varepsilon}(\bm{0}) = D^{\alpha}f^{\varepsilon}(\bm{0})$ for $\forall \left|\alpha\right|\geq 0$. In view of \textbf{[Locality]} we have $A[\bar{f}^{\varepsilon}](\bm{0})=A[f^{\varepsilon}](\bm{0})$. And considering the continuity of $A$, we can deduce $A[f^{\varepsilon}](\bm{0})$ converges to $A[f](\bm{0})$ in $L^{\infty}$ when $\varepsilon$ goes to $0$. This means $A[f](\bm{0}) \geq A[g](\bm{0})$. By symmetry, we can get $A[f](\bm{0}) \leq A[g](\bm{0})$, which means $A[f](\bm{0}) = A[g](\bm{0})$.

Also, in our proof, $\bm{0}$ can be replaced by any $\vx\in \mathbb R^d$. So the value of $A[f](\vx)$ only depends on $\vx,f,Df,D^2f$. Observe that from \textbf{[Linearity]}, $A[f+C]=A[f]$ for any constant $C$, so $A[f](\vx)$ only depends on $\vx,Df,D^2f$. At last, we prove that there exists a continuous function $F$ such that
\[A[f]=F(Df,D^{2}f,\vx)\]
From \textbf{[Comparison Principle]} of $\mathcal T_t$, we can derive a similar argument for $F$. Let $\bm{A} \succeq \bm{B}$ and set
\[f(\vx)=\left[(\vp, \vx)+\frac{1}{2}(\bm{A} \vx, \vx)\right] w(\vx), \qquad g(\vx)=\left[(\vp, \vx)+\frac{1}{2}(\bm{B} \vx, \vx)\right] w(\vx)\]
Indeed, $f \geq g$ on $\mathbb{R}^{d}$ while $f(\bm{0})=g(\bm{0})$. Using \textbf{[Comparison Principle]},
\begin{eqnarray*}
	&& F(\vp,\bm{A}, \bm{0}) =A[f](\bm{0}) =\lim _{t \rightarrow 0^{+}}\frac{T_{t}(f)(\bm{0})-f(\bm{0})}{t} \\
	& \geq &\lim _{t \rightarrow 0^{+}}\frac{T_{t}(g)(\bm{0})-g(\bm{0})}{t}= A[g](\bm{0}) = F(\vp,\bm{B}, \bm{0})
\end{eqnarray*}
$\bm{0}$ can be replaced by any $\vx\in \mathbb R^d$. Thus
\begin{equation}
	\label{eq:a1_compare}
    F(\vp,\bm{A},\vx)\geq F(\vp,\bm{B},\vx) \qquad \text{for any} \quad \bm{A} \succeq \bm{B}
\end{equation}

In the same way, we can get
\[A_{t}[f]=F(D (\mathcal T_t(f)),D^{2} (\mathcal T_t(f)),\vx,t)\]
which implies $u(\vx,t)=\mathcal T_t(f)$ satisfies
\[\begin{cases}
\frac{\partial u(\vx,t)}{\partial t}=F(Du, D^{2}u,\vx,t),\vx\in \mathbb R^{d},t\in [0,T]\\
u(\vx,0)=f(\vx).
\end{cases}\]

According to \textbf{[Linearity]},
$F$ therefore satisfies
\[F(rDf+sDg,rD^{2}f + sD^{2}g,\vx,t) = rF(Df,D^{2}f,\vx,t) + sF(Dg,D^{2}g,\vx,t)\]
for any real numbers $r$ and $s$ and any functions $f$ and $g$ and at any point $(\vx,t)$. Since the values of $Df,Dg,D^2f,D^2g$ are arbitrary and can be independently taken to be 0, we obtain for any vectors $\bm{v}_1,\bm{v}_2$ and symmetric
matrices $\bm{A}_1,\bm{A}_2$ and any fixed point $(\vx_0,t_0)$ that
\begin{eqnarray*}
F(r\bm{v}_1+s\bm{v}_2,r\bm{A}_1+s\bm{A}_2,\vx_0,t_0)&=&rF(\bm{v}_1,\bm{A}_1,\vx_0,t_0)+sF(\bm{v}_2,\bm{A}_2,\vx_0,t_0) \\
F(\bm{v}_1,\bm{A}_1,\vx_0,t_0)&=&F(\bm{v}_1,0,\vx_0,t_0)+F(0,\bm{A}_1,\vx_0,t_0).
\end{eqnarray*}
Let
\[F(\bm{v},0,\vx_0,t_0)= F_1(\bm{v},\vx_0,t_0), F(0,\bm{A},\vx_0,t_0)=F_2(\bm{A},\vx_0,t_0)\]
Then $F_1$ and $F_2$ are both linear, i.e., there exists a function $v:\mathbb R^d\times [0,T]\rightarrow \mathbb R^d$ and a function $\sigma:\mathbb R^d\times [0,T]\rightarrow \mathbb R^{d\times d}$ such that,
\begin{eqnarray*}
 F_1(\bm{v},\vx_0,t_0)&=&v(\vx_0,t_0)\cdot\bm{v},\\
 F_2(\bm{A},\vx_0,t_0)&=&\sum_{i,j}\sigma_{i,j}(\vx,t)A_{i,j},
\end{eqnarray*}
where $A_{i,j}$ is the $i,j$-th element of matrix $\bm{A}$ and $\sigma_{i,j}$ is the $i,j$-th element of matrix function $\sigma$.

If we choose $\bm{A}=\bm{\xi}\bm{\xi}^T\succeq \bm{0}$, where $\bm{\xi}=(\xi_1,\cdots,\xi_d)^T$ is a $d$-dimension vector, then according to \eqref{eq:a1_compare},
\[\bm{\xi}^T\sigma(\vx_0,t_0)\bm{\xi}= \sum_{i,j}\sigma_{i,j}(\vx_0,t_0)\xi_i\xi_j = F_2(\bm{A},\vx_0,t_0)\geq F_2(\bm{0},\vx_0,t_0) = 0\]
which implies matrix function $\sigma$ is a positive semi-definite function.

Thus we can finally get there exist a Lipschitz continuous function $v:\mathbb R^d\times [0,T]\rightarrow \mathbb R^d$ and a Lipschitz continuous positive semi-definite function $\sigma:\mathbb R^d\times [0,T]\rightarrow \mathbb R^{d\times d}$ such that $u(\vx,t)=\mathcal T_{t}(f)$ is the solution of the equation
\begin{equation*}
\begin{cases}
\frac{\partial u(\vx,t)}{\partial t}=v(\vx,t)\cdot \nabla u(\vx,t)+\sum_{i,j}\sigma_{i,j}(\vx,t)\frac{\partial^2 u}{\partial x_i\partial x_j} (\vx,t)\\
u(\vx,0)=f(\vx),
\end{cases}
\end{equation*}
where $\sigma_{i,j}(\vx,t)$ is the $i,j$-th element of matrix function $\sigma(\vx,t)$.
\end{proof}

\section{Additional Experiment results}
First we provide a detailed results for graph node classification tasks.
\begin{table}[hbtp]
\centering
\caption{Performance comparison on graph node classification tasks. In each cell, the first row represents average accuracy $\pm$ standard variation, the second row represents training time per epoch / average convergence time per task (seconds), and the third row represents allocated GPU memory. Results marked with $^*$ indicate that they are averaged across 40 random splits with 10 random initialization each, due to excessive time consumption per task.}
\begin{tabular}{ccccc}
	\toprule
	Category & Method & Cora & Citeseer & Pubmed \\
	\midrule
	\multirow{18}{*}{Classic} & MLP & \makecell{57.4 $\pm$ 2.1 \\ 1.896$\times 10^{-3}$ / 0.420 \\ 1141 MB} & \makecell{59.9 $\pm$ 2.2 \\ 2.085$\times 10^{-3}$ / 0.521 \\1195 MB} & \makecell{70.0 $\pm$ 2.0 \\2.591$\times 10^{-3}$ / 0.504 \\ 4107 MB} \\
	\cmidrule{2-5}
	& \makecell{GCN \\ \cite{kipf2017semi}} & \makecell{81.6 $\pm$ 1.1 \\ 3.072$\times 10^{-3}$ / 1.240 \\ 1147 MB} & \makecell{72.1 $\pm$ 1.6 \\ 3.271$\times 10^{-3}$ / 1.152\\1201 MB}& \makecell{79.0 $\pm$ 2.1 \\ 3.614$\times 10^{-3}$ / 1.617 \\ 4113 MB }  \\
	\cmidrule{2-5}
	& \makecell{GraphSAGE \\ \cite{hamilton2017inductive}} & \makecell{79.3 $\pm$ 1.4 \\ 3.380$\times 10^{-3}$ / 1.161 \\ 1263 MB } & \makecell{71.7 $\pm$ 1.6 \\ 3.646$\times 10^{-3}$ / 1.172 \\ 1339 MB }& \makecell{76.1 $\pm$ 2.0 \\ 4.746$\times 10^{-3}$ / 1.660  \\ 4109 MB } \\
	\cmidrule{2-5}
	& \makecell{GAT\\ \cite{velickovic2018graph}} & \makecell{80.8 $\pm$ 1.3 \\ 7.296$\times 10^{-3}$ / 1.397 \\ 1291 MB } & \makecell{71.6 $\pm$ 1.7 \\ 7.641$\times 10^{-3}$ / 1.324 \\ 1381 MB} & \makecell{78.7 $\pm$ 2.1 \\ 0.0157 / 3.374 \\ 4123 MB } \\
	\cmidrule{2-5}
	& \makecell{SGC\\ \cite{wu2019simplifying}} & \makecell{80.5 $\pm$ 1.3 \\ 1.671$\times 10^{-3}$ / 2.797 \\ 413 MB } & \makecell{73.9 $\pm$ 1.4 \\ 1.531$\times 10^{-3}$ / 2.126 \\ 495 MB } & \makecell{77.2 $\pm$ 2.6 \\ 2.090$\times 10^{-3}$ / 1.545 \\ 527 MB }\\
	\cmidrule{2-5}
	& \makecell{APPNP\\ \cite{gasteiger2019predict}} & \makecell{\textbf{82.7} $\pm$ 1.1 \\ 0.0143 / 12.41 \\ 373 MB } & \makecell{73.3 $\pm$ 1.5 \\ 0.0150 / 8.824 \\ 433 MB} & \makecell{80.6 $\pm$ 1.8 \\ 0.0195 / 8.974 \\ 475 MB }\\
	\midrule
	\multirow{4}{*}{ODE} & \makecell{CGNN\\ \cite{xhonneux2020continuous}} & \makecell{ 82.5$^*$  $\pm$ 1.0\\ 0.3369 / 293.6 \\ 451 MB} & \makecell{73.0$^*$  $\pm$ 1.6 \\ 0.2744 / 110.7 \\ 459 MB} & \makecell{80.5$^*$  $\pm$ 2.2 \\ 0.2678 / 133.2 \\ 951 MB }\\
	\cmidrule{2-5}
	& \makecell{GCDE\\ \cite{poli2021graph}} & \makecell{80.0 $\pm$ 1.5 \\ 0.0205 / 12.24 \\ 387 MB} & \makecell{72.1 $\pm$ 1.6 \\ 0.0207 / 15.53 \\ 429 MB} & \makecell{76.0 $\pm$ 3.9 \\ 0.0216 / 6.309 \\ 3353 MB} \\
	\midrule
	\multirow{18}{*}{Diffusion} & \makecell{GDC\\ \cite{gasteiger2019diffusion}} & \makecell{81.6 $\pm$ 1.3 \\ 8.374$\times 10^{-3}$ / 2.641 \\ 1815 MB} & \makecell{72.2 $\pm$ 2.6 \\ 5.716$\times 10^{-3}$ / 2.421 \\ 1241 MB} & \makecell{79.0 $\pm$ 2.0 \\ 0.0489 / 7.112 \\ 4909 MB } \\
	\cmidrule{2-5}
	& \makecell{GraphHeat\\ \cite{xu2019graph}} & \makecell{81.4 $\pm$ 1.2 \\ 7.803$\times 10^{-3}$ / 3.930 \\ 431 MB} & \makecell{73.5 $\pm$ 1.5 \\ 8.397$\times 10^{-3}$ / 3.256 \\ 467 MB} & \makecell{78.4$^*$  $\pm$ 2.1 \\ 0.2488 / 84.01 \\ 3095 MB} \\
	\cmidrule{2-5}
	& \makecell{DGC\\ \cite{wang2021dissecting}} & \makecell{81.4 $\pm$ 1.2 \\ 1.678$\times 10^{-3}$ / 4.164 \\ 413 MB} & \makecell{75.0 $\pm$ 1.9 \\ 1.797$\times 10^{-3}$ / 0.5149 \\ 495 MB} & \makecell{78.2 $\pm$ 2.1 \\ 4.324$\times 10^{-3}$/ 3.545 \\ 567 MB} \\
	\cmidrule{2-5}
	& \makecell{Difformer\\ \cite{wu2023difformer}} & \makecell{82.0 $\pm$ 2.3 \\ 0.0431 / 18.61 \\ 1205 MB} & \makecell{71.9 $\pm$ 1.7 \\ 0.0265 / 31.27 \\ 1187 MB} & \makecell{74.8 $\pm$ 4.5 \\ 0.0469 / 48.31 \\ 4153 MB } \\
	\cmidrule{2-5}
	& \makecell{GRAND\\ \cite{chamberlain2021grand}} & \makecell{82.5 $\pm$ 1.4 \\ 0.0879 / 11.20 \\ 2879 MB} & \makecell{73.7 $\pm$ 1.7 \\ 0.1257 / 19.39 \\ 2617 MB} & \makecell{78.8 $\pm$ 1.8 \\ 0.3631 / 59.41 \\ 14247 MB}\\
	\cmidrule{2-5}
	& \makecell{Diff-ResNet\\ \cite{wang2024diffusion}} & \makecell{82.1 $\pm$ 1.1 \\ 0.0413 / 12.60 \\ 552 MB} & \makecell{74.6 $\pm$ 1.8 \\ 0.0758 / 16.38 \\ 1163 MB } & \makecell{80.1 $\pm$ 2.0 \\ 0.0509 / 17.35 \\ 829 MB } \\
	\midrule
	& COIN & \makecell{82.2 $\pm$ 1.2 \\ 6.256 $\times 10^{-3}$ / 0.798 \\ 484 MB} & \makecell{\textbf{75.8} $\pm$ 1.3 \\ 0.0134 / 2.081 \\ 906 MB } & \makecell{\textbf{81.1} $\pm$ 1.9 \\ 0.0229 / 3.769 \\ 558 MB} \\
	\bottomrule
\end{tabular}
\end{table}

\newpage
Then we provide a comparison between the reported results (if available) and our re-implemented results of several methods. There are several reasons for the performance discrepancies. Firstly, some methods use a fixed train-val-test split, such as GCDE, GraphHeat, DGC, etc. These methods are likely to overfit on a specific split. Secondly, while some other methods, such as GDC, APPNP, GRAND, also use 100 random train-val-test splits with 20 random neural network initializations each, their validation set has 1500 data points, which is much larger than ours (30 × class number). Lastly, these methods may set a fixed training epoch or utilize various early-stopping criteria, while we use the same early-stopping criterion across all methods in the re-implementation, which is the validation loss does not decrease or validation accuracy does not increase for 50 epochs.

\begin{table}[hbtp]
	\centering
    \caption{Performance difference between reported results (first row) and our re-implementation results (second row).}
	\begin{tabular}{ccccc}
		\toprule
		Category & Method & Cora & Citeseer & Pubmed \\
		\midrule
		\multirow{8}{*}{Classic} & \multirow{2}{*}{GCN} & 80.1 $\pm$ 0.5 & 67.9 $\pm$ 0.5 & 78.9 $\pm$ 0.7\\
        & & 81.6 $\pm$ 1.1 & 72.1 $\pm$ 1.6 & 79.0 $\pm$ 2.1\\
        \cmidrule{2-5}	
		& \multirow{2}{*}{GAT} & 83.0 $\pm$ 0.7 & 72.5 $\pm$ 0.7 & 79.0 $\pm$ 0.3 \\
        & & 80.8 $\pm$ 1.3 & 71.6 $\pm$ 1.7 & 78.7 $\pm$ 2.1 \\
		\cmidrule{2-5}
		& \multirow{2}{*}{SGC} & 80.6 $\pm$ 1.2 & 71.4 $\pm$ 4.0 & 77.0 $\pm$ 1.6\\
		& & 80.5 $\pm$ 1.3 & 73.9 $\pm$ 1.4 & 77.2 $\pm$ 2.6\\
		\cmidrule{2-5}
		& \multirow{2}{*}{APPNP} & 83.8 $\pm$ 0.2 & 75.8 $\pm$ 0.3 & 79.7 $\pm$ 0.3 \\
		& & 82.7 $\pm$ 1.1 & 73.3 $\pm$ 1.5 & 80.6 $\pm$ 1.8\\
		\midrule
		\multirow{4}{*}{ODE} & \multirow{2}{*}{CGNN}& 82.7 $\pm$ 1.2 & 72.7 $\pm$ 0.9 & 83.2 $\pm$ 1.4  \\
		& & 82.5 $\pm$ 1.0 & 73.0 $\pm$ 1.6 & 80.5 $\pm$ 2.2 \\
		\cmidrule{2-5}
		& \multirow{2}{*}{GCDE} & 83.8 $\pm$ 0.5 & 72.5 $\pm$ 0.5 & 79.5 $\pm$ 0.4 \\
		& & 80.0 $\pm$ 1.5 & 72.1 $\pm$ 1.6 & 76.0 $\pm$ 3.9 \\
		\midrule
		\multirow{11}{*}{Diffusion} & \multirow{2}{*}{GDC} & 83.5 $\pm$ 0.2 & 73.2 $\pm$ 0.3 & 79.6 $\pm$ 0.4 \\
		& & 81.6 $\pm$ 1.3 & 72.2 $\pm$ 2.6 & 79.0 $\pm$ 2.0 \\
		\cmidrule{2-5}
		& \multirow{2}{*}{GraphHeat} & 83.7 & 72.5 & 80.5 \\
		& & 81.4 $\pm$ 1.2 & 73.5 $\pm$ 1.5 & 78.4 $\pm$ 2.1 \\
		\cmidrule{2-5}
		& \multirow{2}{*}{DGC} & 83.3 $\pm$ 0.0 & 73.3 $\pm$ 0.1 & 80.3 $\pm$ 0.1 \\
		& & 81.4 $\pm$ 1.2 & 75.0 $\pm$ 1.9 & 78.2 $\pm$ 2.1 \\
		\cmidrule{2-5}
		& \multirow{2}{*}{Difformer} & 85.9 $\pm$ 0.4 & 73.5 $\pm$ 0.3 & 81.8 $\pm$ 0.3 \\
		& & 82.0 $\pm$ 2.3 & 71.9 $\pm$ 1.7 & 74.8 $\pm$ 4.5 \\
		\cmidrule{2-5}
		& \multirow{2}{*}{GRAND} & 83.6 $\pm$ 1.0 & 73.4 $\pm$ 0.5 & 78.8 $\pm$ 1.7\\
		& & 82.5 $\pm$ 1.4 & 73.7 $\pm$ 1.7 & 78.8 $\pm$ 1.8\\
		\bottomrule
	\end{tabular}
\end{table}

\newpage
Table below provides the 1-shot and 5-shot results of few-shot learning with either ResNet-18~(and its variants) or WRN as backbone.

\begin{table}[hbtp]
\footnotesize
\caption{Average accuracy (in \%) and 95\% confidence interval in \textit{mini}ImageNet, \textit{tiered}ImageNet and CUB. Re-implemented results using public official code with our pretrained backbone are marked with \dag.}
\begin{center}
\resizebox{\columnwidth}{!}{\begin{tabular}{lc|cc|cc|cc}
\toprule
& &\multicolumn{2}{c|}{\textbf{\textit{mini}ImageNet}}&\multicolumn{2}{c|}{\textbf{\textit{tiered}ImageNet}}&\multicolumn{2}{c}{\textbf{CUB}} \\
\cline{3-8}
\textbf{Methods} & \textbf{Backbone}&\textbf{1-shot}& \textbf{5-shot}& \textbf{1-shot}& \textbf{5-shot}& \textbf{1-shot}& \textbf{5-shot}\\
\hline
MAML \cite{finn2017model} & ResNet-18 & 49.61 $\pm$ 0.92 & 65.72 $\pm$ 0.77&-&-& 69.96 $\pm$ 1.01&82.70 $\pm$ 0.65\\
Baseline \cite{chen2019closerfewshot} & ResNet-18 & 51.87 $\pm$ 0.77 & 75.68 $\pm$ 0.63&-&-&67.02 $\pm$ 0.90 & 83.58 $\pm$ 0.54\\
RelationNet \cite{sung2018learning} & ResNet-18 & 52.48 $\pm$ 0.86 & 69.83 $\pm$ 0.68& 54.48 $\pm$ 0.93&71.32 $\pm$ 0.78& 67.59 $\pm$ 1.02 & 82.75 $\pm$ 0.58\\
MatchingNet \cite{vinyals2016matching} & ResNet-18 & 52.91 $\pm$ 0.88 & 68.88 $\pm$ 0.69&-&-&72.36 $\pm$ 0.90&83.64 $\pm$ 0.60\\
ProtoNet \cite{snell2017prototypical} & ResNet-18 & 54.16 $\pm$ 0.82 & 73.68 $\pm$ 0.65&53.31 $\pm$ 0.89&72.69 $\pm$ 0.74&71.88 $\pm$ 0.91& 87.42 $\pm$ 0.48\\
Gidaris \cite{gidaris2018dynamic}  & ResNet-15 & 55.45 $\pm$ 0.89 & 70.13 $\pm$ 0.68&-&-&-&-\\
SNAIL \cite{mishra2018simple} & ResNet-15 & 55.71 $\pm$ 0.99 & 68.88 $\pm$ 0.92&-&-&-&-\\
TADAM \cite{oreshkin2018tadam}  & ResNet-15 & 58.50 $\pm$ 0.30 & 76.70 $\pm$ 0.30&-&-&-&-\\
Transductive \cite{dhillon2019baseline} & ResNet-12 & 62.35 $\pm$ 0.66 & 74.53 $\pm$ 0.54&-&-&-&-\\
MetaoptNet \cite{lee2019meta}  & ResNet-18 & 62.64 $\pm$ 0.61 & 78.63 $\pm$ 0.46&65.99 $\pm$ 0.72 & 81.56 $\pm$ 0.53&-&-\\
TPN \cite{liu2018learning} & ResNet-12 & 53.75 $\pm$ 0.86 & 69.43 $\pm$ 0.67 & 57.53 $\pm$ 0.96 & 72.85 $\pm$ 0.74 & - & - \\
TEAM \cite{qiao2019transductive} & ResNet-18 & 60.07 $\pm$ 0.59 & 75.90 $\pm$ 0.38 & - & - & 80.16 $\pm$ 0.52 & 87.17 $\pm$ 0.39 \\
CAN+T \cite{hou2019cross} & ResNet-12 & 67.19 $\pm$ 0.55 & 80.64 $\pm$ 0.35& 73.21 $\pm$ 0.58 & 84.93 $\pm$ 0.38&-&-\\
SimpleShot \cite{wang2019simpleshot}$^{\dag}$& ResNet-18 & 62.86 $\pm$ 0.20 & 79.22 $\pm$ 0.14 & 69.71 $\pm$ 0.23 & 84.13 $\pm$ 0.17 & 72.86 $\pm$ 0.20 & 88.57 $\pm$ 0.11\\
LaplacianShot \cite{ziko2020laplacian}$^{\dag}$ & ResNet-18 & 70.46 $\pm$ 0.23 & 81.76 $\pm$ 0.14 & 76.90 $\pm$ 0.25 & 85.10 $\pm$ 0.17 & 82.92 $\pm$ 0.21 & 90.77 $\pm$ 0.11 \\
EPNet \cite{rodriguez2020embedding}$^{\dag}$ & ResNet-18 & 63.83 $\pm$ 0.20 & 77.98 $\pm$ 0.15 & 70.08 $\pm$ 0.23 & 82.11 $\pm$ 0.18 & 73.32 $\pm$ 0.21 & 87.55 $\pm$ 0.13\\
Diff-ResNet \cite{wang2024diffusion}$^{\dag}$ & ResNet-18 & 71.11 $\pm$ 0.24 & 82.07 $\pm$ 0.14 & 77.98 $\pm$ 0.25 & 85.75 $\pm$ 0.17 & 84.20 $\pm$ 0.21 & 91.12 $\pm$ 0.10\\
COIN & ResNet-18 & \textbf{72.50} $\pm$ 0.24 & \textbf{82.07} $\pm$ 0.14 & \textbf{78.83} $\pm$ 0.25 & \textbf{86.05} $\pm$ 0.16 & \textbf{85.63} $\pm$ 0.20  &   \textbf{91.31} $\pm$ 0.10 \\
\hline
Qiao \cite{qiao2018few} & WRN & 59.60 $\pm$ 0.41 & 73.74 $\pm$ 0.19&-&-&-&-\\
LEO \cite{rusu2018metalearning} & WRN & 61.76 $\pm$ 0.08 &77.59 $\pm$ 0.12& 66.33 $\pm$ 0.05 & 81.44 $\pm$ 0.09&-&-\\
ProtoNet \cite{snell2017prototypical} & WRN & 62.60 $\pm$ 0.20 & 79.97 $\pm$ 0.14&-&-&-&-\\
CC+rot \cite{gidaris2019boosting}  & WRN & 62.93 $\pm$ 0.45 & 79.87 $\pm$ 0.33& 70.53 $\pm$ 0.51 & 84.98 $\pm$ 0.36&-&-\\
MatchingNet \cite{vinyals2016matching} & WRN & 64.03 $\pm$ 0.20 & 76.32 $\pm$ 0.16&-&-&-&-\\
FEAT \cite{ye2020fewshot} & WRN & 65.10 $\pm$ 0.20 & 81.11 $\pm$ 0.14& 70.41 $\pm$ 0.23 & 84.38 $\pm$ 0.16&-&-\\
Transductive \cite{dhillon2019baseline} & WRN & 65.73 $\pm$ 0.68 & 78.40 $\pm$ 0.52& 73.34 $\pm$ 0.71 & 85.50 $\pm$ 0.50&-&-\\
BD-CSPN \cite{liu2020prototype} & WRN & 70.31 $\pm$ 0.93 & 81.89 $\pm$ 0.60& 78.74 $\pm$ 0.95 & 86.92 $\pm$ 0.63&-&-\\
PT+NCM \cite{hu2021leveraging} & WRN & 65.35 $\pm$ 0.20 & \textbf{83.87} $\pm$ 0.13 & 69.96 $\pm$ 0.22 & 86.45 $\pm$ 0.15 & 80.57 $\pm$ 0.20 & 91.15 $\pm$ 0.10\\
SimpleShot \cite{wang2019simpleshot}$^{\dag}$ & WRN & 65.20 $\pm$ 0.20 & 81.28 $\pm$ 0.14 & 71.49 $\pm$ 0.23 & 85.51 $\pm$ 0.16 & 78.62 $\pm$ 0.19 & 91.21 $\pm$ 0.10 \\
LaplacianShot\cite{ziko2020laplacian}$^{\dag}$ & WRN & 72.90 $\pm$ 0.23 & 83.47 $\pm$ 0.14 & 78.79 $\pm$ 0.25 & 86.46 $\pm$ 0.17 & 87.70 $\pm$ 0.18 & 92.73 $\pm$ 0.10\\
EPNet\cite{rodriguez2020embedding}$^{\dag}$ &  WRN & 67.09 $\pm$ 0.21 & 80.71 $\pm$ 0.14 & 73.20 $\pm$ 0.23 & 84.20 $\pm$ 0.17 & 80.88 $\pm$ 0.20 & 91.40 $\pm$ 0.11\\
Diff-ResNet\cite{wang2024diffusion}$^{\dag}$ & WRN & 73.47 $\pm$ 0.23 & 83.86 $\pm$ 0.14 & 79.74 $\pm$ 0.25 & 87.10 $\pm$ 0.16 & 87.74 $\pm$ 0.19 & 92.96 $\pm$ 0.09\\
COIN & WRN & \textbf{74.85} $\pm$ 0.24 & 83.82 $\pm$ 0.14 & \textbf{80.66} $\pm$ 0.25 & \textbf{87.48} $\pm$ 0.15 & \textbf{89.20} $\pm$ 0.18 & \textbf{93.24} $\pm$ 0.09 \\
\bottomrule
\end{tabular}}
\end{center}
\end{table}

Finally, we provide the time complexity results on Cora and Citeseer.

\begin{figure}
    \centering
    \includegraphics[width=0.5\linewidth]{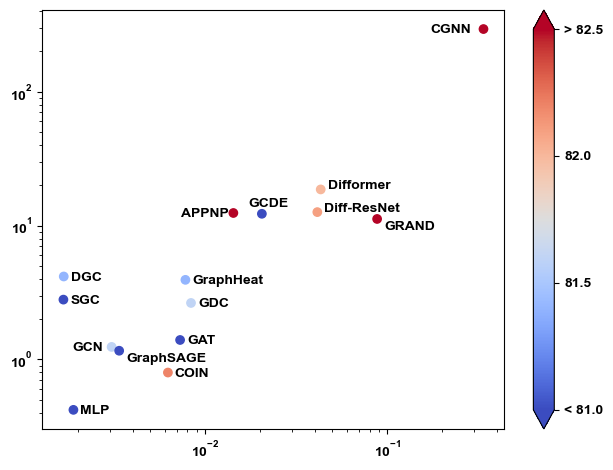}
    \caption{Time complexity on Cora dataset. The x-axis represents average time (seconds) per training epoch, and the y-axis represents average convergence time (seconds) per task, both in log-scale. The color bar measures the average accuracy of each method.}
\end{figure}

\begin{figure}
    \centering
    \includegraphics[width=0.5\linewidth]{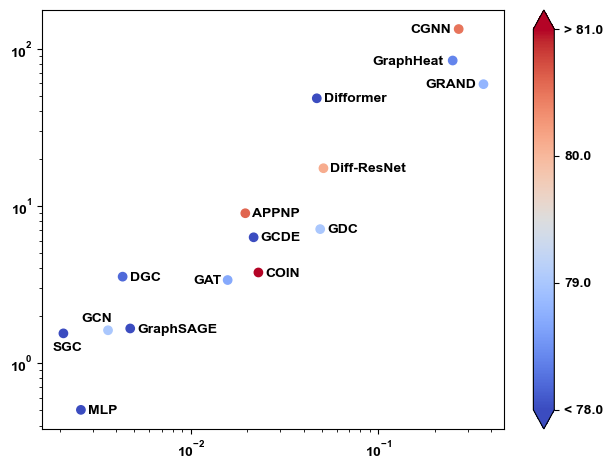}
    \caption{Time complexity on Pubmed dataset. The x-axis represents average time (seconds) per training epoch, and the y-axis represents average convergence time (seconds) per task, both in log-scale. The color bar measures the average accuracy of each method.}
\end{figure}